\pdfoutput=1
\documentclass[11pt]{article}

\usepackage[final]{acl}

\usepackage{times}
\usepackage{latexsym}

\usepackage[T1]{fontenc}
\usepackage[utf8]{inputenc}

\usepackage{microtype}

\usepackage{inconsolata}

\usepackage{graphicx}

\usepackage{multirow}
\usepackage[normalem]{ulem}
\useunder{\uline}{\ul}{}
\usepackage{times}
\usepackage{latexsym}
\usepackage{algorithm}
\usepackage{algpseudocode}
\usepackage{enumitem}
\usepackage[T1]{fontenc}
\usepackage[utf8]{inputenc}
\usepackage{microtype}
\usepackage{inconsolata}
\usepackage{graphicx}
\usepackage{colortbl}
\usepackage{xcolor}         % colors
\usepackage[utf8]{inputenc} % allow utf-8 input
\usepackage[T1]{fontenc}    % use 8-bit T1 fonts
\usepackage{hyperref}       % hyperlinks
\usepackage{url}            % simple URL typesetting
\usepackage{booktabs}       % professional-quality tables
\usepackage{amsfonts}       % blackboard math symbols
\usepackage{nicefrac}       % compact symbols for 1/2, etc.
\usepackage{microtype}      % microtypography
\usepackage{color}
\usepackage{amsmath}
\usepackage{pifont}

\title{
Guided by the Plan: Enhancing Faithful Autoregressive Text-to-Audio Generation with Guided Decoding
}

\author{
 \textbf{Juncheng Wang}$^{1}$\qquad
 \textbf{Zhe Hu}$^{1}$\qquad
 \textbf{Chao Xu}$^{2,3}$\qquad
 \textbf{Siyue Ren}$^{4}$\qquad
\\
 \textbf{Yuxiang Feng}$^{5}$\qquad
\textbf{Yang Liu}$^{2,3}$\qquad
 \textbf{Baigui Sun}$^{2,3}$$^*$\qquad
 \textbf{Shujun Wang}$^{1}$$^*$
\\[0.7em]
 $^1$ The Hong Kong Polytechnic University\quad 
 $^2$ IROOTECH TECHNOLOGY\quad
 $^3$ Wolf 1069 b Lab, Sany Group,\\
 $^4$ Shanghai Artificial Intelligence Laboratory \qquad
 $^5$ Zhejiang University \qquad
}

\begin{document}

\makeatletter
\newcommand\whline{\noalign{\ifnum0=`}\fi\hrule \@height 1.25pt \futurelet
	\reserved@a\@xhline}
\maketitle
\begin{abstract}
Autoregressive (AR) models excel at generating temporally coherent audio by producing tokens sequentially, yet they often falter in faithfully following complex textual prompts—especially those describing complex sound events. 
We uncover a surprising capability in AR audio generators: their early prefix tokens implicitly encode global semantic attributes of the final output, such as event count and sound-object category, revealing a form of \textit{implicit planning}. 
Building on this insight, we propose \textbf{Plan-Critic}, a lightweight auxiliary model trained with a Generalized Advantage Estimation (GAE)-inspired objective to predict final instruction-following quality from partial generations. 
At inference time, Plan-Critic enables \textit{guided exploration}: it evaluates candidate prefixes early, prunes low-fidelity trajectories, and reallocates computation to high-potential planning seeds. 
Our Plan-Critic-guided sampling achieves up to a \textbf{10 points improvement in CLAP score} over the AR baseline—establishing a new state of the art in AR text-to-audio generation—while maintaining computational parity with standard best-of-$N$ decoding. This work bridges the gap between causal generation and global semantic alignment, demonstrating that even strictly autoregressive models can plan ahead. Codes will be available at \hyperlink{code}{https://github.com/wjc2830/Siren.git}.
\end{abstract}

\section{Introduction}
% Audio is inherently temporal—a signal that unfolds causally over time. Yet much of text-to-audio generation has relied on non-autoregressive (AR) frameworks like diffusion models, which often neglect this sequential nature. 
% Inspired by the success of AR in large language models (LLMs), recent work is shifting toward token-based AR audio generation.
% These approaches synthesize audio one step at a time, respecting temporal causality.
% Leading this shift is Siren, a pioneering framework that uses a collaborative transformer to generate multi-layer residual quantized tokens, achieving high-fidelity audio with strong temporal coherence.

Text-to-audio generation aims to synthesize audio signals given textual descriptions~\citep{liu2023audioldm,Tango2,xing2024seeing,audioX}. Most existing approaches adopt non-autoregressive (NAR) frameworks, such as diffusion models~\citep{MMAudio,MelQCD}, that operate over continuous latent representations. 
While highly effective, they often neglect the inherently sequential and causal nature of audio, which unfolds temporally in a strictly forward direction. 
To better capture such dynamics, recent works such as Siren~\citep{wang2025language} has revisited AR paradigm~\citep{audiogen,copet2024simple}, generating audio token-by-token. It naturally mirrors the temporal progression of sound and yields superior temporal coherence.

\begin{figure}
    \centering
    \includegraphics[width=0.9\linewidth]{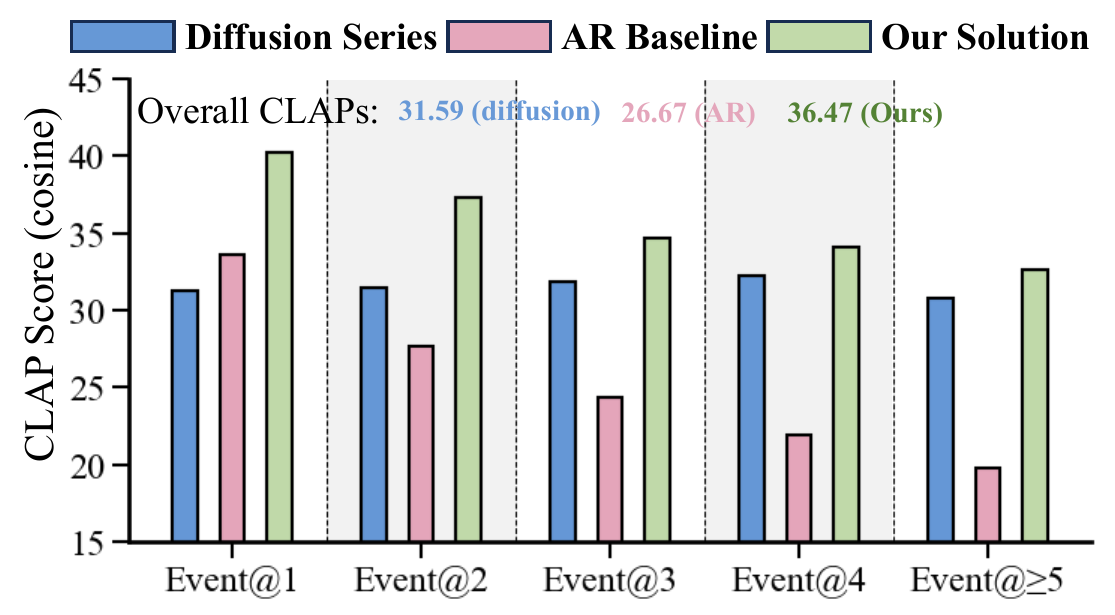}
    \vspace{-2mm}
    \caption{
Instruction-following performance ($y$-axis) of different methods, measured by CLAP scores. The $x$-axis labels, $\text{Event}@N$, indicate prompts with $N$ distinct audio events, where a higher number of events corresponds to increased complexity.    }
\vspace{-4mm}
    \label{fig-teaser}
\end{figure}

Despite their success in producing high-fidelity audio, AR models suffer from a persistent limitation: \textit{they struggle to faithfully follow complex textual prompts}. 
When prompts describe multiple sequential or complex audio events, AR-generated outputs frequently omit or misalign key semantic elements, resulting in degraded instruction-following. 
As shown in Figure~\ref{fig-teaser}, while diffusion models maintain stable performance as prompt complexity increases, AR baselines exhibit a consistent drop in CLAP score~\citep{CLAP}—highlighting a fundamental gap between strong temporal modeling and robust semantic alignment.

The main reason for such disparity lies in architectural differences. 
Diffusion models, often built on UNet~\citep{ldm} or bidirectional transformer~\citep{dit}, allow every token to attend to both past and future states, enabling global coordination and coarse-to-fine planning across the entire timeline. 
AR models, by contrast, generate outputs strictly left-to-right in token level, without explicit global control~\cite{li2022diffusion,hua-wang-2020-pair,shen-etal-2019-towards}. 
As a result, they excel at local temporal coherence but often lack global semantic alignment, a weakness that becomes more pronounced as prompt complexity grows. This raises a central challenge: \textit{how can AR models strengthen instruction following while preserving their causal generation advantages?}

% However, a critical limitation persists in current AR audio generators: \emph{they often fail to faithfully follow complex textual prompts}. When prompts describe multiple concurrent or sequential sound events, generated outputs frequently omit or misalign key components, resulting in degraded semantic fidelity. This shortcoming manifests in lower instruction-following performance compared to diffusion-based counterparts, despite the latter’s weaker temporal continuity.

% A fundamental architectural distinction underlies this disparity.
% Diffusion models typically employ UNet architectures or bidirectional transformers, allowing every audio token to attend with each other—both future and past. 
% This enables explicit coordination of semantic content, facilitating coarse-to-fine planning of the full audio timeline in synthesis. 
% In contrast, AR models generate tokens strictly left-to-right, with no access to future content and only incremental exposure to the prompt through past context. 
% As a result, their decision-making is inherently local and reactive: critical events may be delayed, underrepresented, or entirely missed due to an inability to anticipate long-range dependencies, which can be reflected by the descending performance when the event count in prompt goes larger in Figure.
% This constraint limits global coherence, especially for richly structured prompts—raising a pivotal question:
% \textbf{can we improve instruction following in AR audio generator by incorporating future information grounded in causal generation?}

We draw inspiration from human sequential production, such as speech or writing, where people plan ahead and rely on this plan as a global control during output generation~\cite{hu-etal-2022-planet,hovy1990pragmatics,pan-mckeown-1998-learning-intonation,mckeown1985discourse}. Motivated by this analogy, we first conduct a preliminary analysis with global attribute probing. Our results reveal that AR models exhibit similar signs of implicit planning: the prefix tokens (e.g., the first 32 of a 288-token sequence) already encode predictive information about high-level audio attributes, including the \textit{event count} and the \textit{sound object category}. This finding suggests that AR models, though restricted to causal decoding, implicitly embed global structural cues early in the generation process as \emph{implicit planning}.

% In this work, we draw inspiration from human cognition during speech and writing, where individuals naturally plan ahead while producing sequential output. We investigate whether such a “planning” mechanism exists implicitly within AR audio models like Siren—specifically, whether early-generated partial tokens encode predictive information about global attributes of the complete audio sequence. To probe this, we introduce a novel global attribute probing methodology. We extract two manually designed global attributes from fully synthesized audio: (1) \emph{event count}, the count event extracted from generated audio; and (2) \emph{audio category}, the category of the sounding object. 
% Remarkably, we find that these attributes can be predicted with significant accuracy from information of only the first 32 tokens—what we term the prefix—of a 288-token sequence. This demonstrates that early-generation tokens carry non-trivial signals about the overall structure of future output, suggesting the presence of \textbf{implicit planning} in AR models.

Building on this observation, we propose a guided decoding method that leverages implicit planning as a control signal to steer AR model generation \underline{at inference time}. Specifically, we introduce \textbf{Plan-Critic}, an auxiliary model trained to predict instruction-following quality (measured by CLAP score) from partial audio sequences. In this way, we transfer the implicit planning into a rectified score that enables early selection and guidance of the subsequent full sequence generation. A key challenge in training such critic is credit assignment: the model must infer the overall sequence quality based on partial generations. To address this, we design a \textbf{Generalized Advantage Estimation (GAE)-inspired training framework}: The critic produces value estimates at each generation step, with direct supervision applied only to the final-step prediction using the ground-truth CLAP score. To bridge the gap, intermediate predictions are regularized through temporal consistency losses, propagating supervisory signals backward along the sequence. This strategy mitigates the credit assignment problem and allows reliable assessment during early generation satge.

Equipped with Plan-Critic, we reframe inference as \textit{guided exploration} rather than blind left-to-right rollout. Our \textbf{Plan-Critic-guided sampling} strategy evaluates candidate prefixes early, prunes low-scoring trajectories, and reallocates computation to promising planning seeds. Unlike full-sequence methods like best-of-$N$ sampling—which generate complete outputs before reranking~\cite{stiennon2020learning}—our approach expands the effective search space over high-quality global structures while avoiding redundant computation. This yields more efficient inference-time scaling and significantly stronger alignment with complex prompts.

In summary, our main contributions are:
\begin{itemize}[wide,nosep]
   \item We provide empirical evidence that early tokens in AR audio generation encode predictive information about global semantic attributes, revealing an intrinsic capacity for latent planning.
    \item We propose Plan-Critic—a novel auxiliary model—and a GAE-inspired training paradigm that quantifies implicit planning, enabling accurate prediction of final instruction-following quality from partial sequences.
    \item We introduce a Plan-Critic-guided sampling algorithm that improves both efficiency and semantic fidelity, achieving up to a \textbf{10 points gain in CLAP score} over the AR baseline and establishing a new state-of-the-art in AR text-to-audio generation.
\end{itemize}

\section{On Prefix Tokens and Implicit Planning}
\subsection{Preliminaries of AR Audio Generation}
Consider an audio waveform $x \in \mathbb{R}^{T_{\text{wav}} \times C_{\text{wav}}}$, where $C_{\text{wav}}$ denotes the number of channels and $T_{\text{wav}}$ the temporal duration.  
An AR generator first encodes $x$ into discrete tokens using a residual vector quantization (RVQ) tokenizer~\citep{dac}, yielding a token sequence $\mathbf{q} \in [V]^{R \times T}$, where $V$ is the vocabulary size, $R$ the number of RVQ layers, and $T$ the sequence length in tokens.

Following \citealt{audiogen,wang2025language}, a causal transformer is employed to model the joint distribution over tokens in a left-to-right manner, predicting the next $R$ codes at each time step:
\begin{align}
    &p(\mathbf{q}) 
    = p(q_{1}^{1}, \dots, q_{1}^{R}, \dots, q_{T}^{1}, \dots, q_{T}^{R})  \\
    &= \prod_{t = 1}^{T} p(q_{t}^{1}, \dots, q_{t}^{R} \mid q_{1}^{1}, \dots, q_{1}^{R}, ..., q_{t-1}^{1}, \dots, q_{t-1}^{R}, \Phi(c)),\notag
\end{align}
where $\Phi(c)$ is the encoded textual prompt. After $T$ AR steps, the generated token sequence $\mathbf{q}$ is decoded back into a waveform via a detokenizer.

\subsection{Implicit Planning for Future Generation in Prefix Audio Tokens}
\label{subsec:implicit_planning}
\begin{figure}
    \centering
    \includegraphics[width=\linewidth]{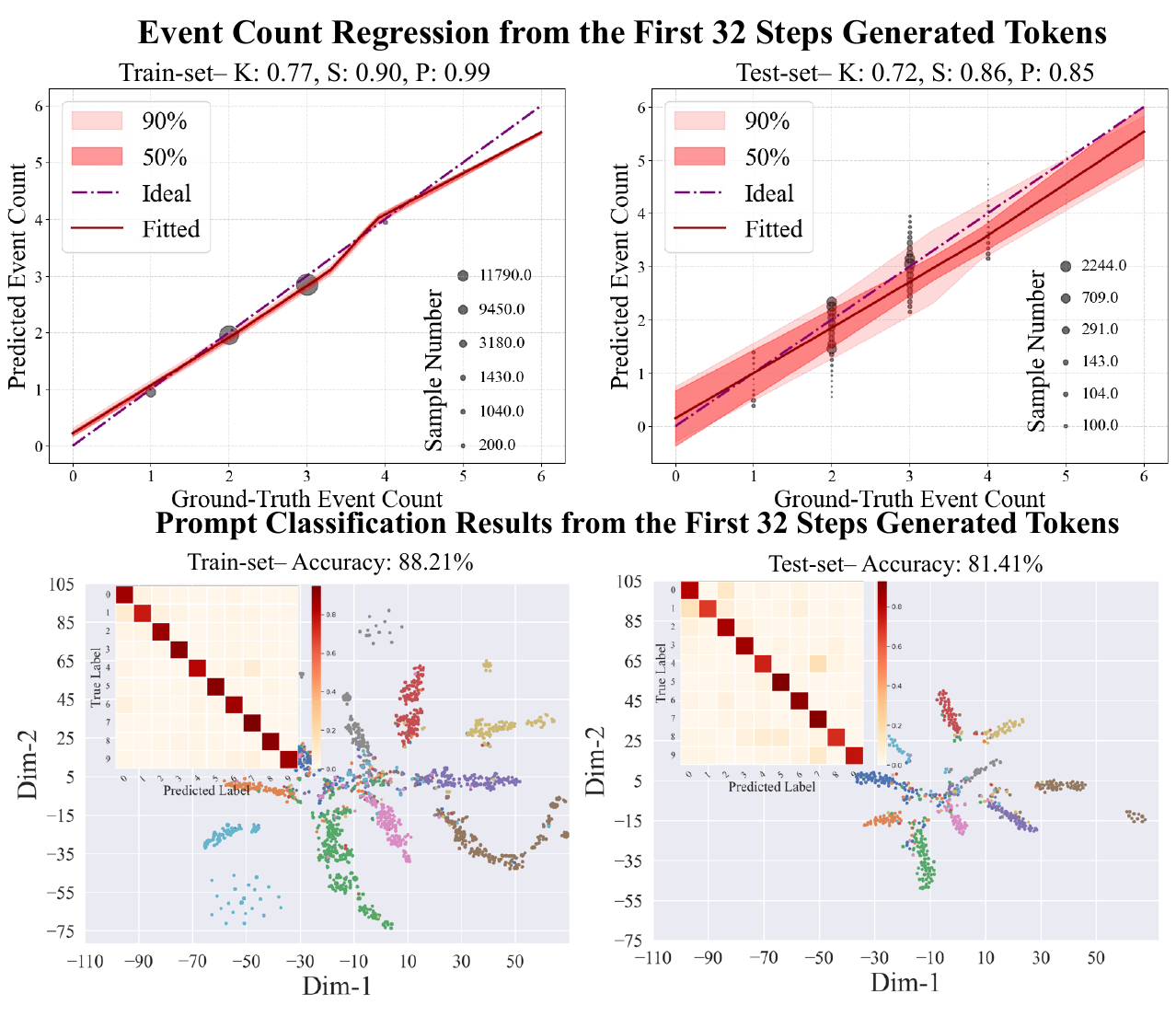}
    \vspace{-6mm}
    \caption{\textbf{Upper row}: Event counting regression results, with Kendall (K), Spearman (S), and Pearson (P) correlation coefficients reported between predicted and ground-truth counts. \textbf{Lower row}: t-SNE visualization of prefix representations, colored by the object category associated with each prefix. The confusion matrix for the classification results is also included. \textbf{Takeaways}: the non-trivial generalization performance of event counting and sounding object classification demonstrate the high correlation between the prefix tokens with posterior global attributes. \emph{Best viewed in color.} }
    \vspace{-4mm}
    \label{fig-findings}
\end{figure}
% Unlike diffusion, which utilizes a bidirectional feature interaction so that can explicitly arrange the whole temporal generation, the partial visibility of AR' s causal order generation results in poor instruction following capacity.
% To remedy this, the similar experiences in diffusion inspire us to incorporate future information in causal decoding to achieve better temporal interaction.

% Motivated by human' s thinking ahead mechanism in writing or speaking, we conduct analytical study to prove there is similar implicit "plan ahead" mechanism in AR generation.
% To prove this, we demonstrate that early generated prefix tokens contains plan ahead information for their entire generation, by showing that some global attributes of their upcoming generation can be predicted from the representations of prefix tokens.
% Following this core idea, we study two correlation probing experiments, where we train a MLP based neural network to model the correlation between the prefix tokens with attributes extracted from its posterior whole generation. With this MLP model achieved non-trivial generalization, it can be inferred that the input (prefix tokens representations) has correlation with output (global attributes extracted from entire audio).

Unlike diffusion-based models~\citep{GenAU,audioX,MMAudio}, which leverage bidirectional context to globally coordinate generation across time, the strictly causal nature of AR models limits their ability to align with complex textual instructions~\citep{hua-wang-2020-pair,shen-etal-2019-towards}. 
This often results in omissions or misalignments of key semantic elements.

Inspired by human cognitive processes in speech and writing, where individuals formulate high-level plans before articulating details~\cite{hu-etal-2022-planet,hovy1990pragmatics}, we hypothesize that AR audio generators similarly engage in \textit{implicit planning}: early generated prefix tokens encode predictive signals about global properties of the full audio sequence.

To test this, we conduct correlation probing experiments: we train lightweight MLPs to predict global attributes of the complete generated audio from representations of its prefix tokens. Successful generalization would imply that such attributes are implicitly encoded during early generation.

% \paragraph{Correlation Probing}
% Given a user provided prompt $\Phi(c)$, we utilize the well-trained AR generator $\pi_{\theta}(\mathbf{q}|\Phi(c))$ to generate whole audio sequences $\mathbf{q}\in[V]^{r\times T}$.
% During the first $T_{prefix}$ temporal steps to generate prefix tokens $\mathbf{q}_{prefix}\in[V]^{R\times T_{prefix}}$, we reserve the penultimate layer features in generator $\pi_{\theta}$ as $\mathbf{h}_{prefix}\in\mathbb{R}^{R\times T_{prefix} \times C}$, where $C$ means hidden dimension. 
% Then, we define the \emph{attribute rule} as $f(\mathbf{q})$, which summarizes the attributes from the whole generated audio (in $T$ temporal length). 

% Intuitively, if the prefix tokens do capture some posterior attributes, we can model the correlation of $g(\mathbf{h}_{prefix})\mapsto f(\mathbf{q})$, where $g$ is a MLP based neural network to fit the correlation. 

\paragraph{Correlation Probing Framework}
With prompt $\Phi(c)$, we use pretrained generator $\pi_\theta(\mathbf{q} \mid \Phi(c))$ to synthesize full audio sequences $\mathbf{q} \in [V]^{R \times T}$. In the first $T_{\text{prefix}}$ steps, we extract the penultimate-layer (for its richest semantics of model) hidden states corresponding to the prefix tokens, $\mathbf{h}_{\text{prefix}} \in \mathbb{R}^{R \times T_{\text{prefix}} \times C}$, with $C$ as hidden dimension.

We define an \emph{attribute function} $f(\mathbf{q})$ that maps the full token sequence to a global semantic attribute (e.g., number of sound events). If $\mathbf{h}_{\text{prefix}}$ contains predictive information about $f(\mathbf{q})$, then a mapping $g(\mathbf{h}_{\text{prefix}}) \mapsto f(\mathbf{q})$—realized by an MLP $g$—should generalize beyond training data.

\paragraph{Attribute Design}
The chosen attributes must be (i) global in scope, (ii) not inferable from local prefix content alone, and (iii) distributed across the full temporal span. 
We select two such attributes that directly influence instruction-following fidelity:
\begin{itemize}[wide]
    \item[1.] \textbf{Event count}: the number of distinct audio events in the generated output. To extract this, we use a multimodal large language model (MLLM)~\citep{comanici2025gemini} to caption the audio and segment the caption into atomic sound events. For example, \texttt{\underline{Motor noise} is followed by a \underline{horn honking} and a \underline{siren wailing}} is parsed into three events, yielding a label of $3$.
        
    \item[2.] \textbf{Sound-producing object category}: the dominant object responsible for sustained sound production. We prompt a LLM~\citep{comanici2025gemini} to generate diverse captions for each of 10 predefined objects (e.g., \texttt{rain}), such as \texttt{Heavy rain is drumming on the window} or \texttt{The rain is tapping quietly on the window during the storm}. All captions derived from the same object share its category label.
\end{itemize}
For attribute (1), the MLP is trained as a regressor; for attribute (2), as a classifier.

% \paragraph{Empirical Results}
% In attribute-1, we randomly generate $5,000$ prompts and let generator $\pi_{\theta}$ to generate multiply audios for each prompt. And we use a MLLM to label the event count of generated audio. 
% With these generated audios, we isolate them into train-, validation-, and test-set. As shown in the upper row of Figure~\ref{fig-findings}, our trained MLP regressor succeeds in generalizing to the test-set with a non-trivial regression performance. 
% Furthermore, given regressed and ground-truth event count, we further utilize Kendall, Spearman, and Pearson to validate their correlation, which all reflects strong correlation.

% In attribute-2, we generate $10$ objects, and let LLM to expand each object with additional $2,000$ prompts. The dataset is also isolated for rigorous evaluation.
% As shown in the lower-half of Figure~\ref{fig-findings}, our trained MLP classifier is able to precisely distinguish the prefix token features.

% \textbf{In conclusion}, the high performance indicates that the high correlation between prefix tokens with global attributes. This correlation present the AR generator is implicitly planning the future generation during generating the prefix tokens.

\paragraph{Empirical Results}
For event count prediction, we generate 5,000 diverse prompts and synthesize multiple audio samples per prompt using $\pi_\theta$. An MLLM labels each sample with its event count. The dataset is split into train, validation, and test sets. As shown in the upper panel of Figure~\ref{fig-findings}, the MLP regressor achieves strong generalization on the test set. Correlation metrics—Kendall, Spearman, and Pearson—all indicate significant alignment between predicted and ground-truth counts.

For object category classification, we generate 2,000 prompts per object across 10 categories. The MLP classifier, trained on prefix representations, achieves high accuracy in distinguishing object categories, as visualized in the lower panel of Figure~\ref{fig-findings}.

\textbf{In summary}, these results demonstrate that prefix tokens in AR audio generation encode rich predictive signals about global semantic attributes of the full sequence. This provides empirical evidence for \textit{implicit planning}—a latent capacity to anticipate future content during early decoding stages.
% \section{Plan-Critic Guided Generation}
% \input{Figures/Fig_Framework}
% Motivated by our above findings, it makes sense that we can preview or anticipate the generated audio' s attributes with only the few start steps.
% And these global attributes (such as event count, audio category) determine the instruction following performance.
% Under this circumstance, we propose to enhance AR' s instruction following performance without retraining its large model.
% To this end, we propose a plan-critic model that is able to transfer the implicit planning performance within prefix tokens into rectified scores that indicates the instruction following performance.
% So that we can utilize this rectified score to critic the prefix token' s planning results.
% Then, with this plan-critic guidance, we can scale up the prefix generation by searching better plans to generate better audios.
\section{Plan-Critic Guided Generation}
\begin{figure}
    \centering
    \includegraphics[width=\linewidth]{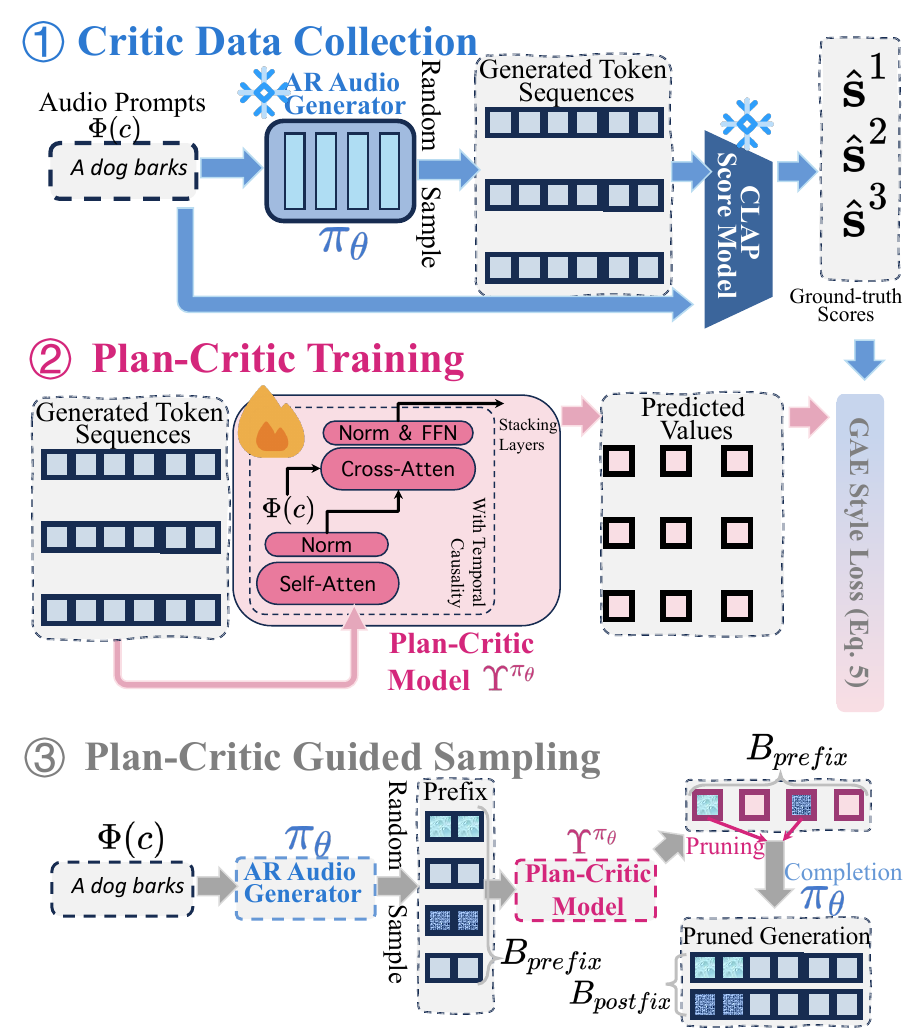}
    \caption{Overall pipeline of the proposed plan-critic guided framework.
\textcircled{1} illustrates the process of collecting training data for the plan-critic model.
\textcircled{2} shows the training of the plan-critic model, where we conduct sparse GAE supervision with certain step $t$ is involved in loss (elaborate in Section~\ref{subsec:setting}).
\textcircled{3} depicts how the trained critic guides the generator’s inference sampling. }
    \label{fig-main_framework}
\end{figure}
Our above insight suggests a practical strategy: \emph{rather than retraining the large AR generator, we can enhance its alignment with complex prompts by leveraging early-generation signals as a form of implicit plan}. To this end, we introduce \textbf{Plan-Critic}, a lightweight auxiliary model that maps prefix tokens to a scalar score estimating the eventual instruction-following quality of the full audio. 
This score enables early evaluation and selection of high-potential generation trajectories, guiding inference toward semantically faithful outputs, with Figure~\ref{fig-main_framework} depicting pipeline.

\subsection{Plan-Critic Model}
The Plan-Critic model, denoted $\Upsilon^{\pi_\theta}$, is a transformer-based critic initialized from a subset of parameters of the pre-trained AR generator $\pi_\theta$. Its purpose is to evaluate, at intermediate generation step $t$, how well the generated audios will perform on instruction from current partial sequence.

\paragraph{Architecture and Initialization}
Given a prefix token sequence $\mathbf{q}_{\text{prefix}} \in [V]^{R \times T_{\text{prefix}}}$, we first embed it into $\mathbf{e}_{\text{prefix}} \in \mathbb{R}^{R \times T_{\text{prefix}} \times C}$ using the embedding layer. Following standard practice~\citep{copet2024simple,wang2025language}, we collapse the $R$ quantization layers via summation to obtain a single temporal sequence $\mathbf{e}'_{\text{prefix}} \in \mathbb{R}^{T_{\text{prefix}} \times C}$.

The prompt $\Phi(c)$ is encoded as a sequence of contextualized text embeddings. The Plan-Critic then processes $\mathbf{e}'_{\text{prefix}}$ and $\Phi(c)$ through interleaved causal self-attention and cross-attention layers. The self-attention captures temporal dynamics within the audio prefix, while cross-attention models alignment between audio and text.

The resulting contextualized representation is:
\begin{equation}
    \mathbf{h}^{\Upsilon} = \Upsilon_h(\mathbf{e}'_{\text{prefix}}, \Phi(c)), \quad \text{where } \mathbf{h}^{\Upsilon} \in \mathbb{R}^{T_{\text{prefix}} \times C}.
    \label{eq:hidden_state}
\end{equation}

To ensure compatibility with the implicit planning signals identified in Section~\ref{subsec:implicit_planning}—which were from the penultimate layer of $\pi_\theta$—we initialize $\Upsilon^{\pi_\theta}$ by corresponding layers of the generator. This alignment ensures the critic operates in the same representational manifold as the planning cues.

Finally, a MLP regression head $\Upsilon_s$ maps the hidden state at step $t$, $\mathbf{h}^{\Upsilon}_{\le t}$, to a scalar value:
\begin{equation}
    s_t = \Upsilon_s(\mathbf{h}^{\Upsilon}_{\le t}).
\end{equation}

\paragraph{Generalized Advantage Estimation Training}
The goal of Plan-Critic is to predict, from a partial sequence, the eventual instruction-following of the completed audio—quantified by CLAP score. 
A naive approach would assign the final CLAP score uniformly to all intermediate steps. However, it ignores the \textit{temporal credit assignment problem}: each step tokens actually contribute differently to the final performance.
% early decisions may be beneficial even if the final output is poor due to later errors, and vice versa.

To address this, we adopt a training objective inspired by Generalized Advantage Estimation (GAE) from reinforcement learning~\cite{PPO}. For a generated sequence $\mathbf{q}$ of length $T$, the Plan-Critic produces a sequence of value estimates $\{s_t\}_{t=1}^T$. Let $\hat{s}$ denote the ground-truth CLAP score of the full sequence. We define a sparse reward signal $\hat{\mathbf{s}} \in \mathbb{R}^T$ where $\hat{\mathbf{s}}_T = \hat{s}$ and $\hat{\mathbf{s}}_t = 0$ for $t < T$.

The GAE-based target for step $t$ is:
\begin{equation}
    r_t = \sum_{l=0}^{T-t-1} (\gamma \lambda)^l \left( \hat{\mathbf{s}}_{t+l} + \gamma s^{\text{old}}_{t+l+1} - s^{\text{old}}_{t+l} \right) + s^{\text{old}}_t,
\end{equation}
where $s^{\text{old}}$ denotes the value estimates from a detached (non-gradient) copy of the critic, and $\gamma, \lambda \in [0,1]$ are discount and smoothing hyperparameters. The critic is trained to minimize:
\begin{equation}
    \mathcal{L}_{\text{critic}} = \mathbb{E}_t \left[ \frac{1}{2} (s_t - r_t)^2 \right].
    \label{eq:gae}
\end{equation}

\textbf{Remark (Intuition behind Eq.~\ref{eq:gae}).}
The Eq.~\ref{eq:gae} target $r_t$ propagates credit backward from the final reward while discounting distant contributions (via $\gamma$) and smoothing temporal differences (via $\lambda$). This enables the critic to assign higher value to prefixes that genuinely support semantic alignment.
% —even when the final audio is imperfect due to later decoding errors. 

\subsection{Plan-Critic Guided Sampling}

Equipped with a trained Plan-Critic, we reformulate inference as a guided search over generation trajectories. Standard AR sampling performs a blind left-to-right rollout, allocating equal computational resources to all time steps. However, our analysis in Section~\ref{subsec:implicit_planning} shows that the prefix tokens encode a high-level plan that largely determines semantic fidelity. Moreover, empirical results (Section~\ref{subsec:exp_postfix}) confirm that once the prefix is fixed, further exploration in the postfix yields diminishing returns in instruction-following performance.

This motivates a \textit{prefix-first} inference strategy: allocate more sampling budget to exploring diverse prefix candidates, evaluate them early using the Plan-Critic, and prune low-scoring trajectories before committing to full-sequence generation.

Our guided sampling proceeds as follows:
\begin{enumerate}
    \item Given a prompt $\Phi(c)$, sample a large batch of prefixes $\{\mathbf{q}_{\text{prefix}}^i\}_{i=1}^{B_{\text{prefix}}}$ from $\pi_\theta(\cdot \mid \Phi(c))$.
    \item At step $T_{\text{prefix}}$, critic each prefix' s planning to obtain scores $s_{T_{\text{prefix}}}^i = \Upsilon_s(\mathbf{q}_{\text{prefix}}^i, \Phi(c))$.
    \item Select the top-$B_{\text{postfix}}$ prefixes ($B_{\text{postfix}} < B_{\text{prefix}}$) with the highest scores.
    \item Continue AR generation from these selected prefixes to produce full audio sequences.
\end{enumerate}

This approach dynamically reallocates computation toward high-potential planning seeds, improving both efficiency and semantic alignment. In our experiments, we fix the total token budget to match that of the baseline AR model (i.e., $B_{\text{prefix}} \cdot T_{\text{prefix}} + B_{\text{postfix}} \cdot (T - T_{\text{prefix}}) = B_{\text{baseline}} \cdot T$), ensuring a fair comparison. Details on hyperparameter selection are provided in Section~\ref{subsec:ablate}.

\section{Experiment}
\subsection{Settings}
\label{subsec:setting}
\paragraph{Datasets}
To train Plan-Critic model on base generator unseen prompts, we use a large language model (LLM) to synthesize 5,000 pseudo audio captions and prompt our baseline generator, Siren~\citep{wang2025language}, to produce 32 audio samples per caption.

To mitigate potential domain shift from LLM-generated prompts, we additionally curate prompts from the official test sets of AudioCaps~\citep{kim2019audiocaps} and VGGSound~\citep{chen2020vggsound}—ensuring they were not seen during Siren’s original training. From each dataset, we reserve 1,000 such prompts as held-out evaluation environments for assessing final performance.

\paragraph{Training Details}
% Siren’s official implementation employs six parallel Transformers, each responsible for two layers of residual vector quantization (RVQ) codes. Directly initializing the plan-critic from the full Siren model would be computationally prohibitive. Guided by Siren’s own analysis—which shows that the first Transformer (handling the top two RVQ layers) carries the majority of semantic information—we initialize our plan-critic exclusively from this first Transformer.

To enable effective credit assignment across long sequences, we adopt a sparse supervision strategy for the Generalized Advantage Estimation (GAE) objective: the critic outputs scores only at every 32nd time step, i.e., for all t such that $mod(t,32)=0$. This ensures gradient signals from the final reward can propagate meaningfully back to the prefix tokens. Moreover, to avoid overfitting, we randomly add a Gaussian noise upon the computed CLAP scores.

The plan-critic is trained using the AdamW optimizer with a learning rate of $1e-4$ for $40k$ steps.

\paragraph{Inference Details}
During inference, we employ a prefix-first search strategy: we generate $B_{prefix}=128$ candidate prefixes in parallel and retain only the top $B_{postfix}=2$ based on plan-critic scores for full-sequence completion. The prefix length is fixed at $T_{prefix}=32$ tokens.

For a full audio sequence of $288$ tokens per RVQ layer (with 12 RVQ layers total), our method 

\begin{table*}[t]
\centering

\renewcommand{\arraystretch}{1.15}{
\resizebox{\textwidth}{!}{
\begin{tabular}{ccccccccccc}
\whline
\multicolumn{1}{c|}{\multirow{2}{*}{\textbf{Method}}} & \multicolumn{6}{c|}{\textbf{Instruction Following $\uparrow$}}                                                                                                                                                                           & \multicolumn{4}{c}{\textbf{Audio Quality}}                                                                                                   \\ \cline{2-11} 
\multicolumn{1}{c|}{}                                 & \multicolumn{1}{c|}{\textbf{Overall CLAP}} & \multicolumn{1}{c|}{CLAP@1}         & \multicolumn{1}{c|}{CLAP@2}         & \multicolumn{1}{c|}{CLAP@3}         & \multicolumn{1}{c|}{CLAP@4}         & \multicolumn{1}{c|}{CLAP@$\ge$5}        & \multicolumn{1}{c|}{FAD$\downarrow$} & \multicolumn{1}{c|}{FD$\downarrow$} & \multicolumn{1}{c|}{IS$\uparrow$} & KL$\downarrow$ \\ \whline
\multicolumn{11}{c}{Solutions with Bidirectional Generators}                                                                                                                                                                                                                                                                                                                                                                                    \\ \hline
\multicolumn{1}{c|}{AudioLDM2}                        & \multicolumn{1}{c|}{24.46}                 & \multicolumn{1}{c|}{30.89}          & \multicolumn{1}{c|}{24.75}          & \multicolumn{1}{c|}{22.12}          & \multicolumn{1}{c|}{21.93}          & \multicolumn{1}{c|}{20.30}          & \multicolumn{1}{c|}{2.04}            & \multicolumn{1}{c|}{37.76}          & \multicolumn{1}{c|}{8.32}                      & 1.72            \\ \hline
\multicolumn{1}{c|}{MagNet}                           & \multicolumn{1}{c|}{22.04}                 & \multicolumn{1}{c|}{19.68}          & \multicolumn{1}{c|}{22.51}          & \multicolumn{1}{c|}{22.22}          & \multicolumn{1}{c|}{24.51}          & \multicolumn{1}{c|}{19.26}          & \multicolumn{1}{c|}{4.64}            & \multicolumn{1}{c|}{33.24}          & \multicolumn{1}{c|}{8.20}                      & 1.99            \\ \hline
\multicolumn{1}{c|}{AudioX}                           & \multicolumn{1}{c|}{27.12}                 & \multicolumn{1}{c|}{27.14}          & \multicolumn{1}{c|}{27.19}          & \multicolumn{1}{c|}{27.11}          & \multicolumn{1}{c|}{28.49}          & \multicolumn{1}{c|}{25.36}          & \multicolumn{1}{c|}{3.02}            & \multicolumn{1}{c|}{28.68}          & \multicolumn{1}{c|}{11.28}                     & 1.54            \\ \hline
\multicolumn{1}{c|}{MMAudio}                           & \multicolumn{1}{c|}{31.21}                 & \multicolumn{1}{c|}{32.94}          & \multicolumn{1}{c|}{31.13}          & \multicolumn{1}{c|}{31.13}          & \multicolumn{1}{c|}{30.78}          & \multicolumn{1}{c|}{28.17}          & \multicolumn{1}{c|}{5.51}            & \multicolumn{1}{c|}{19.06}          & \multicolumn{1}{c|}{13.78}                     & 1.37            \\ \hline
\multicolumn{1}{c|}{TangoFlux}                        & \multicolumn{1}{c|}{33.43}                 & \multicolumn{1}{c|}{31.36}          & \multicolumn{1}{c|}{32.94}          & \multicolumn{1}{c|}{34.68}          & \multicolumn{1}{c|}{\underline{ 34.28}}    & \multicolumn{1}{c|}{\textbf{35.05}} & \multicolumn{1}{c|}{2.89}            & \multicolumn{1}{c|}{25.54}          & \multicolumn{1}{c|}{\underline{ 13.75}}               & \textbf{1.16}   \\ \hline
\multicolumn{1}{c|}{GenAU}                            & \multicolumn{1}{c|}{\underline{ 34.59}}           & \multicolumn{1}{c|}{\underline{ 33.68}}    & \multicolumn{1}{c|}{\underline{ 34.57}}    & \multicolumn{1}{c|}{\underline{ 34.54}}    & \multicolumn{1}{c|}{\textbf{35.36}} & \multicolumn{1}{c|}{\underline{ 34.59}}    & \multicolumn{1}{c|}{\textbf{1.73}}   & \multicolumn{1}{c|}{23.01}          & \multicolumn{1}{c|}{13.63}                     & 1.28            \\ \hline
\multicolumn{11}{c}{Solutions with Unidirectional Generators}                                                                                                                                                                                                                                                                                                                                                                                   \\ \hline
\multicolumn{1}{c|}{DelayPattern+BoN-16}                     & \multicolumn{1}{c|}{20.01}                 & \multicolumn{1}{c|}{21.26}          & \multicolumn{1}{c|}{20.27}          & \multicolumn{1}{c|}{19.44}          & \multicolumn{1}{c|}{19.76}          & \multicolumn{1}{c|}{17.76}          & \multicolumn{1}{c|}{3.58}            & \multicolumn{1}{c|}{17.82}          & \multicolumn{1}{c|}{10.67}                     & 2.08            \\ \hline
\multicolumn{1}{c|}{Siren+BoN-16}                            & \multicolumn{1}{c|}{26.67}                 & \multicolumn{1}{c|}{33.63}          & \multicolumn{1}{c|}{27.68}          & \multicolumn{1}{c|}{24.38}          & \multicolumn{1}{c|}{21.95}          & \multicolumn{1}{c|}{19.82}          & \multicolumn{1}{c|}{\underline{ 1.95}}      & \multicolumn{1}{c|}{\underline{ 17.63}}    & \multicolumn{1}{c|}{10.41}                     & 1.91            \\ \hline
\multicolumn{1}{c|}{Siren+Ours}                       & \multicolumn{1}{c|}{\textbf{36.47}}        & \multicolumn{1}{c|}{\textbf{40.20}} & \multicolumn{1}{c|}{\textbf{37.35}} & \multicolumn{1}{c|}{\textbf{34.72}} & \multicolumn{1}{c|}{34.09}          & \multicolumn{1}{c|}{32.59}          & \multicolumn{1}{c|}{1.96}            & \multicolumn{1}{c|}{\textbf{15.70}} & \multicolumn{1}{c|}{\textbf{13.82}}            & \underline{ 1.27}      \\ \whline
\end{tabular}
}}
\caption{Main results on AudioCaps. The best performed metric is in \textbf{bold}, and the second best is \underline{underlined}. Among solutions with bidirectional solutions, MagNet also utilizes a RVQ discrete tokenizer, but is equipped with a BERT-like transformer in its generator. For unidirectional solutions, DelayPattern and Siren are all equipped with inference decoding method of best-of-16. For Siren and ours solution, we share the same generated token budget, where for the first policy transformer, the generated token numbers are all 4,608. }
\label{tab-main_result_AC}
\end{table*}
\noindent consumes $128\times32+2\times(288-32)=4,608$ tokens per RVQ layer, matching the token budget of Siren' s official best-of-\emph{N} (BoN-16) baseline ($16\times288=4,608$). For notational simplicity, we omit the factor of 12 (RVQ layers) in subsequent discussions, reporting token counts per layer.

All experiments are conducted on a single-node AMD MI300X server.
\paragraph{Metrics}
Following prior work~\citep{MMAudio,audioX,wang2025language}, we evaluate audio quality using Fréchet Audio Distance (FAD), Fréchet Distance (FD), Kullback–Leibler (KL) divergence, and Inception Score (IS).
And CLAP for overall instruction following.
To be fine grained, we assess model behavior across complexity, where we use LLM to classify test prompts by the number of audio events and report performance across different event counts.

\subsection{Main Results}
\paragraph{Comparison with Other Audio Generators}
In Table~\ref{tab-main_result_AC} and \ref{tab-main_result_VGGSound}, we compare our method with some bidirectional solutions, including AudioLDM2~\citep{liu2024audioldm}, MagNet~\citep{ziv2024masked}, AudioX~\citep{audioX}, MMAudio~\citep{MMAudio}, TangoFlux~\citep{hung2024tangoflux}, and GenAU~\citep{GenAU}. As for unidirectional solutions, we pick DelayPattern~\citep{copet2024simple} and our baseline Siren~\citep{wang2025language}.
As the tables shown, our method (Siren+Ours) achieves consistently non-trivial improvement over baseline Siren.
This demonstrates that critic-guided prefix search enables superior instruction alignment.

\paragraph{Comparison with Other Advanced Decoding Algorithms}
With plan-critic model trained as an instant reward model for ARs, it can be incorporated with existing guided decoding methods, including BoN, importance sampling~\citep{owen2000safe} and TreeBoN~\citep{qiu2024treebon}.
Results in Table~\ref{tab-compareTTS_AC} and \ref{tab-compareTTS_VGG} show our presented sampling best cooperates with our plan-critic model as the best.

% \usepackage[normalem]{ulem}
% \useunder\underline{ine}\underline{}{}
\begin{table}[]
\centering

\renewcommand{\arraystretch}{1.15}{
\resizebox{.5\textwidth}{!}{
\begin{tabular}{c|c|c|c|c|c}
\whline
\textbf{Method}     & \textbf{CLAP}$\uparrow$  & FAD$\downarrow$ & FD$\downarrow$ & IS$\uparrow$   & KL$\downarrow$ \\ \hline
BoN                 & 26.67          & \textbf{1.95}   & {\ul 17.63}    & {\ul 10.41}    & 1.91           \\ \hline
Importance Sampling & 29.72          & 2.52            & 27.58          & 9.91           & 2.90           \\ \hline
TreeBoN             & {\ul 34.04}    & 1.97            & 18.26          & 10.10          & {\ul 1.38}     \\ \hline
\rowcolor[HTML]{EFEFEF} 
Ours                & \textbf{36.47} & {\ul 1.96}      & \textbf{15.70} & \textbf{13.82} & \textbf{1.27}  \\ \whline
\end{tabular}
}}
\caption{Comparison with other advanced decoding algorithms on AudioCaps, where these methods share the same our trained plan-critic model.}
\label{tab-compareTTS_AC}
\end{table}
% \usepackage[normalem]{ulem}
% \useunder\underline{ine}\underline{}{}
\begin{table}[]
\centering

\renewcommand{\arraystretch}{1.15}{
\resizebox{.5\textwidth}{!}{
\begin{tabular}{c|c|c|c|c|c}
\whline
\textbf{Method}     & \textbf{CLAP$\uparrow$}  & FAD$\downarrow$ & FD$\downarrow$ & IS$\uparrow$   & KL$\downarrow$ \\ \whline
BoN                 & 25.62          & 4.63            & 32.23          & 9.56           & 3.02           \\ \hline
Importance Sampling & 27.37          & 4.40            & 30.84          & 9.19           & {\ul 3.01}     \\ \hline
TreeBoN             & {\ul 30.90}    & \textbf{3.64}   & {\ul 29.41}    & {\ul 10.01}    & 3.33           \\ \hline
\rowcolor[HTML]{EFEFEF} 
Ours                & \textbf{35.88} & {\ul 3.70}      & \textbf{28.02} & \textbf{10.65} & \textbf{2.87}  \\ \whline
\end{tabular}
}}
\caption{Comparison with other advanced decoding algorithms on VGGSound.}
\label{tab-compareTTS_VGG}
\end{table}

\subsection{Finding Studies}
\label{subsec:exp_postfix}
\begin{figure}
    \centering
    \includegraphics[width=0.9\linewidth]{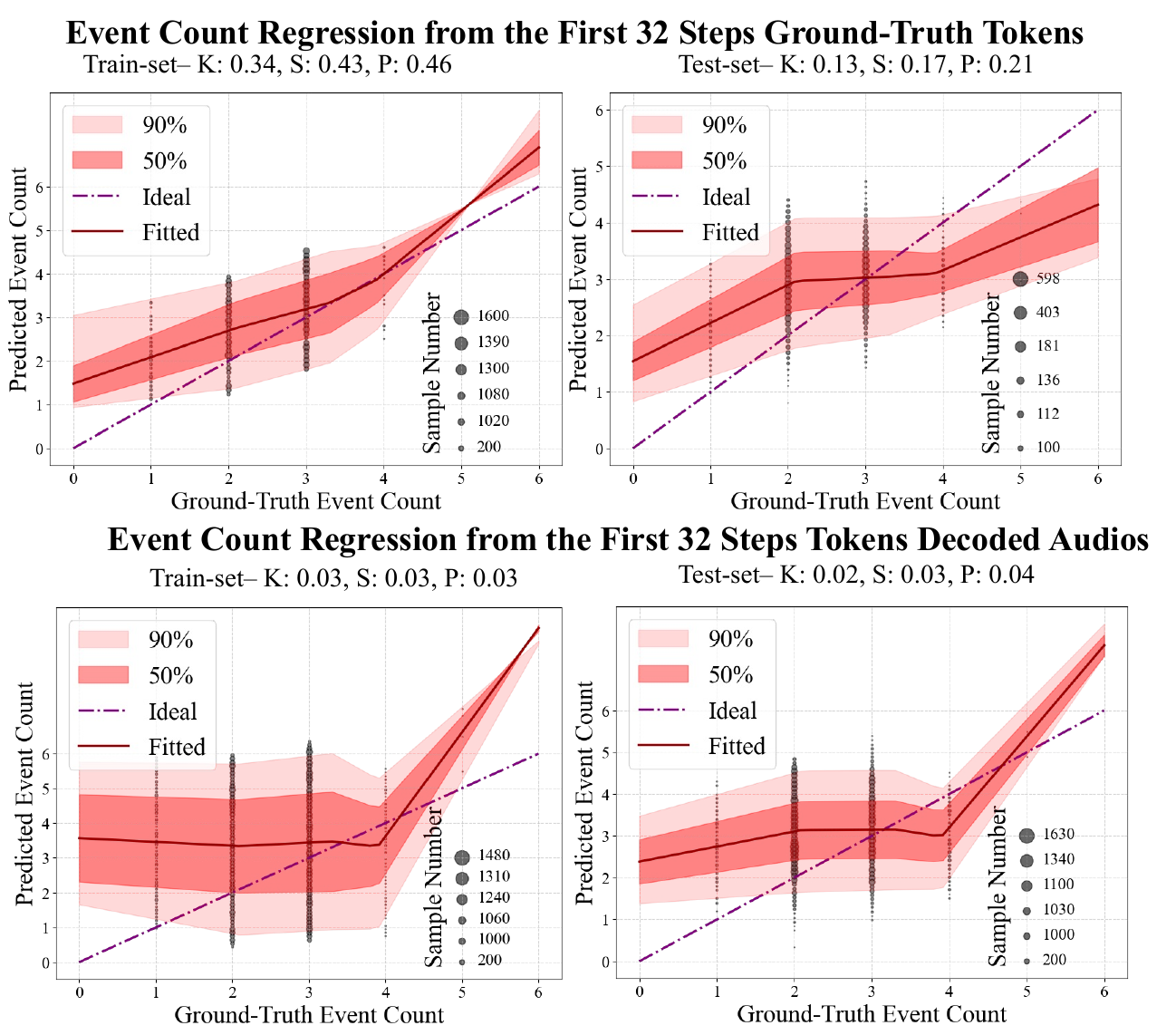}
    \caption{\textbf{Upper row}: 
    Using prefix tokens from real audio to regress the total event count of the full audio. \textbf{Low row}: Event count regression using prefix tokens from decoded (reconstructed) audio.}
    \label{fig-findings_support}
\end{figure}

In this section, we conduct findings experiments to support our major insights: the prefix tokens indeed contain predictive information to the global attributes of posteriorly generated audio.
\paragraph{Correlation with attributes is not because prefix tokens decoded audio content}
To support that the correlation obtained in Section~\ref{subsec:implicit_planning} is not because the prefix tokens decoded audios have contained corresponding attributes, we conduct a comparison experiment that we utilize the detokenizer to decode the prefix tokens into audio, 
% \usepackage{multirow}
% \usepackage[normalem]{ulem}
% \useunder\underline{ine}\underline{}{}
\begin{table*}[t]
\centering

\renewcommand{\arraystretch}{1.15}{
\resizebox{\textwidth}{!}{
\begin{tabular}{c|cccccc|cccc}
\whline
\multirow{2}{*}{\textbf{Method}} & \multicolumn{6}{c|}{\textbf{Instruction Following $\uparrow$}}                                                                                                                                                      & \multicolumn{4}{c}{\textbf{Audio Quality}}                                                                                        \\ \cline{2-11} 
                                 & \multicolumn{1}{c|}{\textbf{Overall CLAP}} & \multicolumn{1}{c|}{CLAP@1}         & \multicolumn{1}{c|}{CLAP@2}         & \multicolumn{1}{c|}{CLAP@3}         & \multicolumn{1}{c|}{CLAP@4}         & CLAP@$\ge$5        & \multicolumn{1}{c|}{FAD$\downarrow$} & \multicolumn{1}{c|}{FD$\downarrow$} & \multicolumn{1}{c|}{IS$\uparrow$}   & KL$\downarrow$ \\ \whline
AudioX                           & \multicolumn{1}{c|}{\underline{ 33.93}}           & \multicolumn{1}{c|}{\underline{ 37.71}}    & \multicolumn{1}{c|}{\underline{ 33.60}}    & \multicolumn{1}{c|}{32.32}          & \multicolumn{1}{c|}{\underline{ 29.61}}    & \underline{ 28.52}    & \multicolumn{1}{c|}{4.44}            & \multicolumn{1}{c|}{35.86}          & \multicolumn{1}{c|}{9.39}           & 2.94            \\ \hline
MMAudio                          & \multicolumn{1}{c|}{31.69}                 & \multicolumn{1}{c|}{36.81}          & \multicolumn{1}{c|}{31.39}          & \multicolumn{1}{c|}{29.45}          & \multicolumn{1}{c|}{26.02}          & 22.81          & \multicolumn{1}{c|}{\underline{3.57}}   & \multicolumn{1}{c|}{\underline{ 29.10}}    & \multicolumn{1}{c|}{\underline{ 10.64}}    & 2.90            \\ \hline
GenAU                            & \multicolumn{1}{c|}{33.03}                 & \multicolumn{1}{c|}{36.76}          & \multicolumn{1}{c|}{32.90}          & \multicolumn{1}{c|}{\underline{ 33.60}}    & \multicolumn{1}{c|}{27.04}          & \underline{ 21.10}    & \multicolumn{1}{c|}{\textbf{3.44}}   & \multicolumn{1}{c|}{34.55}          & \multicolumn{1}{c|}{10.10}          & \textbf{2.52}   \\ \hline
Siren+BoN-16                            & \multicolumn{1}{c|}{25.62}                 & \multicolumn{1}{c|}{28.60}          & \multicolumn{1}{c|}{26.51}          & \multicolumn{1}{c|}{22.51}          & \multicolumn{1}{c|}{21.53}          & 16.37          & \multicolumn{1}{c|}{4.63}            & \multicolumn{1}{c|}{32.23}          & \multicolumn{1}{c|}{9.56}           & 3.02            \\ \hline
Siren+Ours                       & \multicolumn{1}{c|}{\textbf{35.88}}        & \multicolumn{1}{c|}{\textbf{38.83}} & \multicolumn{1}{c|}{\textbf{35.56}} & \multicolumn{1}{c|}{\textbf{35.53}} & \multicolumn{1}{c|}{\textbf{32.86}} & \textbf{28.86} & \multicolumn{1}{c|}{3.70}            & \multicolumn{1}{c|}{\textbf{28.02}} & \multicolumn{1}{c|}{\textbf{10.65}} & \underline{ 2.87}      \\ \whline
\end{tabular}
}}
\caption{Main results on VGGSound, where the textual prompts are annotated by Gemini-2.5 pro, and further polished with a CLAP score restriction with original audio.}
\vspace{-4mm}
\label{tab-main_result_VGGSound}
\end{table*}
and replace the MLP network with a VGGish audio feature extraction model, and try to regress the event count.
As shown in the lower row of Figure~\ref{fig-findings_support}, we fail to build this correlation from audio, which reflects the correlation is not because audio content.

Moreover, we conduct a comparison, that given some ground-truth audios collected from real world, we utilize Siren' s tokenizer~\citep{dac} to transfer them into token sequences, where we find that the first $32$-step tokens fail to regress event counts, which further demonstrates \textbf{implicit-planning} is the property of AR generators.

\paragraph{Postfix tokens yields diminishing influence in the instruction following}
\begin{figure}
    \centering
    \includegraphics[width=\linewidth]{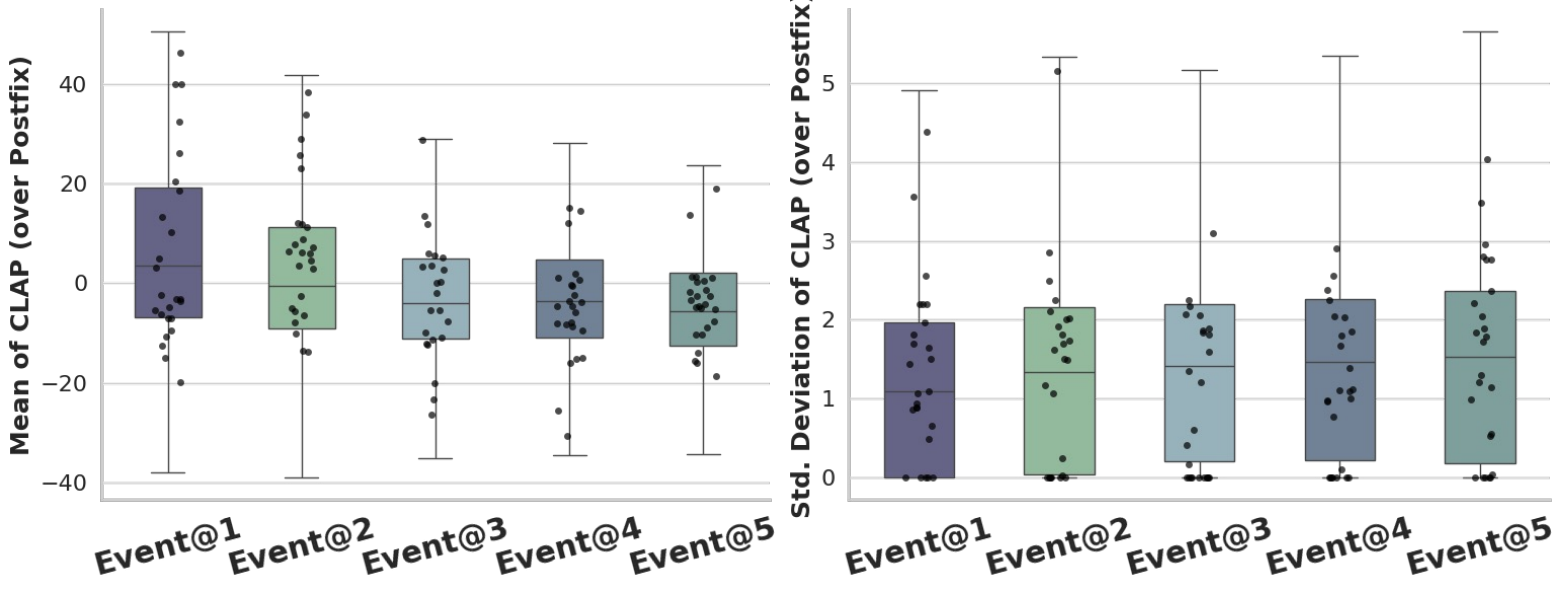}
    \caption{Given a set of prefix tokens, each is expanded into 1,000 postfix sequences. The two plots show the distributions of the \textbf{mean} and \textbf{standard deviation} of CLAP scores computed across the 1,000 generations per prefix, aggregated over 10,000 distinct prefixes. To simplify the figure, we randomly select and visualize 25 points for each event count.}
    \label{fig-postfix}
\end{figure}
We investigate how much instruction-following quality is determined by prefix tokens. For 10,000 fixed prefixes, we generate 1,000 full audio completions each and compute the mean and standard deviation of the CLAP scores over 1,000 post completions. 
In Figure~\ref{fig-postfix}, the means $[-10,20]$, which means significant prefix influences. While standard deviations are tightly clustered in $[2,3]$. This narrow variance indicates that postfix variations have minimal impact.

\subsection{Ablation Studies}
In this section, we conduct extensive ablation study over the choices of hyper-parameters involved in our method. 
In the following tables, we highlight the rows with our final option in gray.
\label{subsec:ablate}
\paragraph{Ablating plan-critic model training}
In Table~\ref{tab-ablate_GAE}, we compare our plan-critic model training' s signal with a naive sparse signal (tokens within a sequence shares the same CLAP score). As a comparison, we compare it with a random guess method (using the same inference sampling strategy expect critic-then-prune). 
In Table~\ref{tab-ablate_gae_interval}, we ablate different choices of GAE supervision interval steps.

% \usepackage[normalem]{ulem}
% \useunder\underline{ine}\underline{}{}
\begin{table}[]
\centering

\renewcommand{\arraystretch}{1.15}{
\resizebox{.45\textwidth}{!}{
\begin{tabular}{c|c|c|c|c|c}
\whline
\textbf{Components} & \textbf{CLAP$\uparrow$} & FAD$\downarrow$ & FD$\downarrow$ & IS$\uparrow$ & KL$\downarrow$ \\ \whline
Baseline            & 26.67                   & 1.95            & 17.63          & 10.41        & 1.91           \\ \hline
Sparse Supervision  & 10.67                   & 3.01            & 37.66          & 8.16         & 3.52           \\ \hline
Random Guess        & 12.86                   & 2.47            & 23.19          & 8.50         & 3.09           \\ \hline
\rowcolor[HTML]{EFEFEF} Ours                & 36.47                   & 1.96            & 15.70          & 13.82        & 1.27           \\ \whline
\end{tabular}
}}
\caption{Ablating plan-critic model training' s supervision signal.}
\vspace{-3mm}
\label{tab-ablate_GAE}
\end{table}
% \usepackage[normalem]{ulem}
% \useunder\underline{ine}\underline{}{}
\begin{table}[]
\centering

\renewcommand{\arraystretch}{1.15}{
\resizebox{.5\textwidth}{!}{

\begin{tabular}{c|c|c|c|c|c|c}
\whline
\textbf{Sampling Strategy} & \textbf{Cost Tokens$\downarrow$} & \textbf{CLAP$\uparrow$} & FAD$\downarrow$ & FD$\downarrow$ & IS$\uparrow$ & KL$\downarrow$ \\ \whline
Baseline                   & 4608                             & 26.67                   & 1.95            & 17.63          & 10.41        & 1.91           \\ \hline
Interval-16                & 2592                             & 31.27                   & 1.90            & 17.02          & 12.57        & 1.64           \\ \hline
\rowcolor[HTML]{EFEFEF} 
Interval-32                & 4608                             & 36.47                   & 1.96            & 15.70          & 13.82        & 1.27           \\ \hline
Interval-64                & 8640                             & 37.34                   & 1.76            & 15.21          & 13.48        & 1.21           \\ \whline
\end{tabular}

}}
\caption{Ablating the interval step of GAE training.}
\label{tab-ablate_gae_interval}
\end{table}

\paragraph{Ablating sampling strategies}
In Table~\ref{tab-ablate_critic_step}, we ablate different choices for the timing of Plan-Critic evaluation, specifically the prefix length $T_{\text{prefix}}$ at which the critic is applied. To clarify it: \textit{Prefix-$x$} denotes applying critic after the first $x$ temporal steps, while \textit{Postfix-$x$} is applying it $x$ steps before the end of sequence (i.e., at step $288 - x$). 

To ensure that the \textit{Postfix-32} configuration can be executed within the memory constraints of a single GPU, we use a reduced prefix batch size of $B_{\text{prefix}} = 8$ uniformly across all entries in Table~\ref{tab-ablate_critic_step}. 

Additional ablation studies exploring alternative sampling strategies are provided in the Appendix.
% \usepackage[normalem]{ulem}
% \useunder\underline{ine}\underline{}{}
\begin{table}[]
\centering

\renewcommand{\arraystretch}{1.15}{
\resizebox{.4\textwidth}{!}{

\begin{tabular}{c|c|c|c|c|c}
\whline
\textbf{Critic Time} & \textbf{CLAP$\uparrow$} & FAD$\downarrow$ & FD$\downarrow$ & IS$\uparrow$ & KL$\downarrow$ \\ \whline
\rowcolor[HTML]{EFEFEF} 
Prefix-32            & 27.52                   & 1.86   & 18.07          & 11.12        & 1.85           \\ \hline
Prefix-64            & 18.81                   & 2.40            & 21.84          & 8.88         & 2.64           \\ \hline
Prefix-96            & 18.56                   & 2.43            & 22.19          & 8.88         & 2.68           \\ \hline
Postfix-64          & 18.40                   & 2.55            & 22.31          & 8.86         & 2.69           \\ \hline
Postfix-32           & 18.01                   & 2.50            & 21.91          & 9.01         & 2.71           \\ \whline
\end{tabular}

}}
\caption{Ablating the critic temporal step.}
\label{tab-ablate_critic_step}
\end{table}

\subsection{User Study}
% \usepackage[normalem]{ulem}
% \useunder\underline{ine}\underline{}{}
\begin{table}[]
\centering

\renewcommand{\arraystretch}{1.15}{
\resizebox{.4\textwidth}{!}{
\begin{tabular}{c|cc|cc}
\whline
\multirow{2}{*}{Method} & \multicolumn{2}{c|}{Instruction Following}      & \multicolumn{2}{c}{Audio Fidelity}    \\ \cline{2-5} 
                        & \multicolumn{1}{c|}{General} & Complex & \multicolumn{1}{c|}{General} & Complex \\ \whline
vs. Siren               & \multicolumn{1}{c|}{78.36}            & 93.20   & \multicolumn{1}{c|}{53.53}   & 55.20   \\ \hline
vs. AudioX              & \multicolumn{1}{c|}{70.49}            & 67.89   & \multicolumn{1}{c|}{60.81}   & 65.37   \\ \hline
vs. GenAU               & \multicolumn{1}{c|}{65.01}            & 63.37   & \multicolumn{1}{c|}{45.61}   & 50.19   \\ \whline
\end{tabular}
}}
\caption{User study results. We report the win-rate (\%) of our solution when comparing with different methods.}
\label{tab-user_study}
\end{table}
In Table~\ref{tab-user_study}, we conducted a human evaluation study with 30 domain experts. Each annotator assessed approximately 100 prompts per semantic category using pairwise A/B tests. For each pair, they judged which audio better followed the prompt instructions and which sounded more natural. This yielded over 18,000 comparisons in total, providing human-validated support for our method’s gains in both semantic fidelity and audio quality.
\section{Related Work}

\noindent\textbf{Audio Generation.}
Denoising diffusion models~\citep{ddpm,scorebased,ddim,adm,dpm-solver,ldm,dit} excel at audio synthesis. 
They are typically applied in latent space: 1D tokenizers operate on waveform~\citep{SAO,ETTA}, while 2D architectures process mel-spectrograms using U-Nets~\citep{liu2022diffsinger,liu2023audioldm,liu2024audioldm,xue2024auffusion,evans2024long,xing2024seeing,du2023conditional,liu2024tell,agostinelli2023musiclm,Tango2,MelQCD} or diffusion transformers (DiT)~\citep{ETTA,Fugatto}. 
Recent work integrates flow matching~\citep{FlowMatching,FMRecon} to accelerate sampling and achieve sota results~\citep{MMAudio,Fugatto}. However, the bidirectional feature interaction breaks the nature of causal flow of audio data.

In parallel, ARs adopt discrete representations via vector quantization (VQ)~\citep{igpt,maskgit,mage,magvit,movq,muse,var}. Early hybrids suffered from fidelity loss due to VQ compression~\citep{magvit2,fsq}, prompting the adoption of residual VQ (RVQ)~\citep{wu2019vector,encodec,dac}. A key advance is Siren~\citep{wang2025language}, which employs a parallel transformer and achieves sota audio quality. 
Yet, it exhibits weak instruction-following capability. 
% To alleviate it, we propose plan-critic guided generation, it significantly improves instruction following without compromising audio fidelity.

\smallskip
\noindent\textbf{Test-Time Compute.}
Our work is also related to recent advances in test-time compute, which aim to improve model performance by generating multiple candidate outputs and then selecting the best one according to specific criteria~\cite{snell2024scaling,ji2025survey,muennighoff2025s1}. Common strategies include majority voting~\cite{wang2022self,toh2024not} and Best-of-$N$ selection based on predefined rewards~\cite{cobbe2021training,levi2024simple}. A central component of this paradigm is the use of reward models to evaluate either the final outcomes~\cite{liu2024acemath,xin2024deepseek} or the intermediate reasoning steps~\cite{ma2023let,khalifa2025process}. In this work, we extend test-time compute to text-to-audio generation by introducing an implicit planning–based partial reward model, enabling effective and efficient scaling of AR models for high-quality audio synthesis.

% \section{Conclusion}
% We have shown that autoregressive text-to-audio models, despite their causal generation constraints, exhibit an intrinsic capacity for \textit{implicit planning}: early prefix tokens encode predictive signals about global semantic attributes of the final output. Leveraging this insight, we proposed \textbf{Plan-Critic}, a lightweight auxiliary model trained with a GAE-inspired objective to estimate instruction-following quality from partial sequences. Coupled with a guided batch-descending sampling strategy, our method reallocates computation toward high-potential planning seeds, significantly improving both semantic fidelity and inference efficiency. Experiments demonstrate up to a \textbf{10\% gain in CLAP score} over the Siren baseline, establishing a new state of the art in AR text-to-audio generation. Our work bridges the gap between local coherence and global alignment, revealing that even strictly autoregressive systems can plan ahead—when guided wisely.
\section{Conclusion}
In this work, we show that AR text-to-audio models inherently perform implicit planning: early tokens encode global semantic attributes of the full output. Based on this, we propose \textbf{Plan-Critic}, a lightweight model trained with a GAE-inspired objective to estimate instruction-following quality from partial sequences. Leveraging Plan-Critic, we further introduce a guided sampling strategy that reallocates computation to high-potential prefixes, yielding up to a \textbf{10\% CLAP score gain} over Siren and setting a new state of the art. Our results demonstrate that even strictly causal models can achieve global alignment—when guided wisely.
\section*{Limitations}
While our Plan-Critic framework significantly improves instruction-following fidelity in autoregressive (AR) text-to-audio generation, several limitations remain:
\paragraph{Dependency on CLAP as a Proxy Metric:}
Our Plan-Critic is trained to predict CLAP scores, which serve as a proxy for semantic alignment. However, CLAP—like any embedding-based metric—may not fully capture nuanced aspects of human perception, such as temporal plausibility, emotional tone, or fine-grained event ordering. 
To remedy this, our future work may expand from designing a human preference aligned reward model.
\paragraph{Fixed Prefix Length Assumption:}
Our method assumes that global planning is encoded within a fixed-length prefix (e.g., the first 32 tokens). While empirical results support this for the evaluated model and audio duration (288 tokens), this may not generalize to longer or more complex audio sequences (e.g., multi-minute soundscapes), where planning might unfold over multiple stages or require hierarchical structure.
To remedy this, the potential future work may explore the prefix length with a fixed ratio with the whole length.
\paragraph{Training Data Bias:}
The Plan-Critic is trained on audio generated by a single AR model (Siren) using LLM-synthesized and curated prompts. This introduces potential biases: the critic may overfit to Siren’s specific failure modes or stylistic tendencies and may not transfer well to other AR generators or domains with different acoustic characteristics (e.g., music vs. environmental sounds).
However, the scaling capacity of transformer architecture along with training data has been well validated. The following works may address this from scaling up the critic model size and training data.

\section*{Acknowledgment}
This work was partially supported by RGC Collaborative Research Fund (No. C5055-24G), the Start-up Fund of The Hong Kong Polytechnic University (No. P0045999), the Seed Fund of the Research Institute for Smart Ageing (No. P0050946), and Tsinghua-PolyU Joint Research Initiative Fund (No. P0056509), and PolyU UGC funding (No. P0053716).

\bibliography{ref}

@string{ICASSP="ICASSP"}

@string{EMNLP="EMNLP"}

@article{qiu2024treebon,
  title={Treebon: Enhancing inference-time alignment with speculative tree-search and best-of-n sampling},
  author={Qiu, Jiahao and Lu, Yifu and Zeng, Yifan and Guo, Jiacheng and Geng, Jiayi and Zhu, Chenhao and Juan, Xinzhe and Yang, Ling and Wang, Huazheng and Huang, Kaixuan and others},
  journal={arXiv preprint arXiv:2410.16033},
  year={2024}
}

@article{owen2000safe,
  title={Safe and effective importance sampling},
  author={Owen, Art and Zhou, Yi},
  journal={Journal of the American Statistical Association},
  volume={95},
  number={449},
  pages={135--143},
  year={2000},
  publisher={Taylor \& Francis}
}

@inproceedings{Fugatto,
  title={Fugatto 1: Foundational Generative Audio Transformer Opus 1},
  author={Valle, Rafael and Badlani, Rohan and Kong, Zhifeng and Lee, Sang-gil and Goel, Arushi and Kim, Sungwon and Santos, Joao Felipe and Dai, Shuqi and Gururani, Siddharth and Aljafari, Aya and others},
  booktitle={International Conference on Learning Representations},
  year={2025}
}

@inproceedings{SAO,
  title={Stable audio open},
  author={Evans, Zach and Parker, Julian D and Carr, CJ and Zukowski, Zack and Taylor, Josiah and Pons, Jordi},
  booktitle={ICASSP 2025-2025 IEEE International Conference on Acoustics, Speech and Signal Processing (ICASSP)},
  year={2025},
}

@inproceedings{Tango2,
  title={Tango 2: Aligning diffusion-based text-to-audio generations through direct preference optimization},
  author={Majumder, Navonil and Hung, Chia-Yu and Ghosal, Deepanway and Hsu, Wei-Ning and Mihalcea, Rada and Poria, Soujanya},
  booktitle={Proceedings of the 32nd ACM International Conference on Multimedia},
  pages={564--572},
  year={2024}
}

@inproceedings{MelQCD,
  title={Synchronized Video-to-Audio Generation via Mel Quantization-Continuum Decomposition},
  author={Wang, Juncheng and Xu, Chao and Yu, Cheng and Shang, Lei and Hu, Zhe and Wang, Shujun and Bo, Liefeng},
  booktitle={Proceedings of the IEEE/CVF conference on computer vision and pattern recognition},
  year={2025}
}

@article{ETTA,
  title={ETTA: Elucidating the Design Space of Text-to-Audio Models},
  author={Lee, Sang-gil and Kong, Zhifeng and Goel, Arushi and Kim, Sungwon and Valle, Rafael and Catanzaro, Bryan},
  journal={arXiv preprint arXiv:2412.19351},
  year={2024}
}

@article{ddpm,
  title={Denoising diffusion probabilistic models},
  author={Ho, Jonathan and Jain, Ajay and Abbeel, Pieter},
  journal={Advances in neural information processing systems},
  volume={33},
  pages={6840--6851},
  year={2020}
}

@article{scorebased,
  title={Generative modeling by estimating gradients of the data distribution},
  author={Song, Yang and Ermon, Stefano},
  journal={Advances in neural information processing systems},
  volume={32},
  year={2019}
}

@article{ddim,
  title={Denoising diffusion implicit models},
  author={Song, Jiaming and Meng, Chenlin and Ermon, Stefano},
  journal={arXiv preprint arXiv:2010.02502},
  year={2020}
}

@article{adm,
  title={Diffusion models beat gans on image synthesis},
  author={Dhariwal, Prafulla and Nichol, Alexander},
  journal={Advances in neural information processing systems},
  volume={34},
  pages={8780--8794},
  year={2021}
}

@article{dpm-solver,
  title={Dpm-solver: A fast ode solver for diffusion probabilistic model sampling in around 10 steps},
  author={Lu, Cheng and Zhou, Yuhao and Bao, Fan and Chen, Jianfei and Li, Chongxuan and Zhu, Jun},
  journal={Advances in Neural Information Processing Systems},
  volume={35},
  pages={5775--5787},
  year={2022}
}

@inproceedings{ldm,
  title={High-resolution image synthesis with latent diffusion models},
  author={Rombach, Robin and Blattmann, Andreas and Lorenz, Dominik and Esser, Patrick and Ommer, Bj{\"o}rn},
  booktitle={Proceedings of the IEEE/CVF conference on computer vision and pattern recognition},
  pages={10684--10695},
  year={2022}
}

@inproceedings{dit,
  title={Scalable diffusion models with transformers},
  author={Peebles, William and Xie, Saining},
  booktitle={Proceedings of the IEEE/CVF International Conference on Computer Vision},
  pages={4195--4205},
  year={2023}
}

@inproceedings{maskgit,
  title={Maskgit: Masked generative image transformer},
  author={Chang, Huiwen and Zhang, Han and Jiang, Lu and Liu, Ce and Freeman, William T},
  booktitle={Proceedings of the IEEE/CVF Conference on Computer Vision and Pattern Recognition},
  pages={11315--11325},
  year={2022}
}

@inproceedings{mage,
  title={Mage: Masked generative encoder to unify representation learning and image synthesis},
  author={Li, Tianhong and Chang, Huiwen and Mishra, Shlok and Zhang, Han and Katabi, Dina and Krishnan, Dilip},
  booktitle={Proceedings of the IEEE/CVF Conference on Computer Vision and Pattern Recognition},
  pages={2142--2152},
  year={2023}
}

@inproceedings{magvit,
  title={Magvit: Masked generative video transformer},
  author={Yu, Lijun and Cheng, Yong and Sohn, Kihyuk and Lezama, Jos{\'e} and Zhang, Han and Chang, Huiwen and Hauptmann, Alexander G and Yang, Ming-Hsuan and Hao, Yuan and Essa, Irfan and others},
  booktitle={Proceedings of the IEEE/CVF Conference on Computer Vision and Pattern Recognition},
  pages={10459--10469},
  year={2023}
}

@article{movq,
  title={Movq: Modulating quantized vectors for high-fidelity image generation},
  author={Zheng, Chuanxia and Vuong, Tung-Long and Cai, Jianfei and Phung, Dinh},
  journal={Advances in Neural Information Processing Systems},
  volume={35},
  pages={23412--23425},
  year={2022}
}

@article{fsq,
  title={Finite scalar quantization: Vq-vae made simple},
  author={Mentzer, Fabian and Minnen, David and Agustsson, Eirikur and Tschannen, Michael},
  journal={arXiv preprint arXiv:2309.15505},
  year={2023}
}

@article{muse,
  title={Muse: Text-to-image generation via masked generative transformers},
  author={Chang, Huiwen and Zhang, Han and Barber, Jarred and Maschinot, AJ and Lezama, Jose and Jiang, Lu and Yang, Ming-Hsuan and Murphy, Kevin and Freeman, William T and Rubinstein, Michael and others},
  journal={arXiv preprint arXiv:2301.00704},
  year={2023}
}

@article{var,
  title={Visual Autoregressive Modeling: Scalable Image Generation via Next-Scale Prediction},
  author={Tian, Keyu and Jiang, Yi and Yuan, Zehuan and Peng, Bingyue and Wang, Liwei},
  journal={arXiv preprint arXiv:2404.02905},
  year={2024}
}

@article{magvit2,
  title={Language Model Beats Diffusion--Tokenizer is Key to Visual Generation},
  author={Yu, Lijun and Lezama, Jos{\'e} and Gundavarapu, Nitesh B and Versari, Luca and Sohn, Kihyuk and Minnen, David and Cheng, Yong and Gupta, Agrim and Gu, Xiuye and Hauptmann, Alexander G and others},
  journal={arXiv preprint arXiv:2310.05737},
  year={2023}
}

@article{hung2024tangoflux,
  title={Tangoflux: Super fast and faithful text to audio generation with flow matching and clap-ranked preference optimization},
  author={Hung, Chia-Yu and Majumder, Navonil and Kong, Zhifeng and Mehrish, Ambuj and Bagherzadeh, Amir Ali and Li, Chuan and Valle, Rafael and Catanzaro, Bryan and Poria, Soujanya},
  journal={arXiv preprint arXiv:2412.21037},
  year={2024}
}

@inproceedings{igpt,
  title={Generative pretraining from pixels},
  author={Chen, Mark and Radford, Alec and Child, Rewon and Wu, Jeffrey and Jun, Heewoo and Luan, David and Sutskever, Ilya},
  booktitle={International conference on machine learning},
  pages={1691--1703},
  year={2020},
  organization={PMLR}
}

@article{liu2024audioldm,
  title={Audioldm 2: Learning holistic audio generation with self-supervised pretraining},
  author={Liu, Haohe and Yuan, Yi and Liu, Xubo and Mei, Xinhao and Kong, Qiuqiang and Tian, Qiao and Wang, Yuping and Wang, Wenwu and Wang, Yuxuan and Plumbley, Mark D},
  journal={IEEE/ACM Transactions on Audio, Speech, and Language Processing},
  year={2024},
  publisher={IEEE}
}

@article{liu2023audioldm,
  title={Audioldm: Text-to-audio generation with latent diffusion models},
  author={Liu, Haohe and Chen, Zehua and Yuan, Yi and Mei, Xinhao and Liu, Xubo and Mandic, Danilo and Wang, Wenwu and Plumbley, Mark D},
  journal={arXiv preprint arXiv:2301.12503},
  year={2023}
}

@inproceedings{liu2022diffsinger,
  title={Diffsinger: Singing voice synthesis via shallow diffusion mechanism},
  author={Liu, Jinglin and Li, Chengxi and Ren, Yi and Chen, Feiyang and Zhao, Zhou},
  booktitle={Proceedings of the AAAI conference on artificial intelligence},
  volume={36},
  number={10},
  pages={11020--11028},
  year={2022}
}

@article{evans2024long,
  title={Long-form music generation with latent diffusion},
  author={Evans, Zach and Parker, Julian D and Carr, CJ and Zukowski, Zack and Taylor, Josiah and Pons, Jordi},
  journal={arXiv preprint arXiv:2404.10301},
  year={2024}
}

@article{ziv2024masked,
  title={Masked audio generation using a single non-autoregressive transformer},
  author={Ziv, Alon and Gat, Itai and Lan, Gael Le and Remez, Tal and Kreuk, Felix and D{\'e}fossez, Alexandre and Copet, Jade and Synnaeve, Gabriel and Adi, Yossi},
  journal={arXiv preprint arXiv:2401.04577},
  year={2024}
}

@inproceedings{xing2024seeing,
  title={Seeing and hearing: Open-domain visual-audio generation with diffusion latent aligners},
  author={Xing, Yazhou and He, Yingqing and Tian, Zeyue and Wang, Xintao and Chen, Qifeng},
  booktitle={Proceedings of the IEEE/CVF Conference on Computer Vision and Pattern Recognition},
  pages={7151--7161},
  year={2024}
}

@inproceedings{du2023conditional,
  title={Conditional generation of audio from video via foley analogies},
  author={Du, Yuexi and Chen, Ziyang and Salamon, Justin and Russell, Bryan and Owens, Andrew},
  booktitle={Proceedings of the IEEE/CVF Conference on Computer Vision and Pattern Recognition},
  pages={2426--2436},
  year={2023}
}

@article{liu2024tell,
  title={Tell What You Hear From What You See--Video to Audio Generation Through Text},
  author={Liu, Xiulong and Su, Kun and Shlizerman, Eli},
  journal={arXiv preprint arXiv:2411.05679},
  year={2024}
}

@article{agostinelli2023musiclm,
  title={Musiclm: Generating music from text},
  author={Agostinelli, Andrea and Denk, Timo I and Borsos, Zal{\'a}n and Engel, Jesse and Verzetti, Mauro and Caillon, Antoine and Huang, Qingqing and Jansen, Aren and Roberts, Adam and Tagliasacchi, Marco and others},
  journal={arXiv preprint arXiv:2301.11325},
  year={2023}
}

@article{xue2024auffusion,
  title={Auffusion: Leveraging the power of diffusion and large language models for text-to-audio generation},
  author={Xue, Jinlong and Deng, Yayue and Gao, Yingming and Li, Ya},
  journal={arXiv preprint arXiv:2401.01044},
  year={2024}
}

@inproceedings{kim2019audiocaps,
  title={Audiocaps: Generating captions for audios in the wild},
  author={Kim, Chris Dongjoo and Kim, Byeongchang and Lee, Hyunmin and Kim, Gunhee},
  booktitle={Proceedings of the 2019 Conference of the North American Chapter of the Association for Computational Linguistics: Human Language Technologies, Volume 1 (Long and Short Papers)},
  pages={119--132},
  year={2019}
}

@inproceedings{chen2020vggsound,
  title={Vggsound: A large-scale audio-visual dataset},
  author={Chen, Honglie and Xie, Weidi and Vedaldi, Andrea and Zisserman, Andrew},
  booktitle={ICASSP 2020-2020 IEEE International Conference on Acoustics, Speech and Signal Processing (ICASSP)},
  pages={721--725},
  year={2020},
  organization={IEEE}
}

@article{copet2024simple,
  title={Simple and controllable music generation},
  author={Copet, Jade and Kreuk, Felix and Gat, Itai and Remez, Tal and Kant, David and Synnaeve, Gabriel and Adi, Yossi and D{\'e}fossez, Alexandre},
  journal={Advances in Neural Information Processing Systems},
  volume={36},
  year={2024}
}

@article{wu2019vector,
  title={Vector quantization: a review},
  author={Wu, Ze-bin and Yu, Jun-qing},
  journal={Frontiers of Information Technology \& Electronic Engineering},
  volume={20},
  number={4},
  pages={507--524},
  year={2019},
  publisher={Springer}
}

@article{encodec,
  title={High fidelity neural audio compression},
  author={D{\'e}fossez, Alexandre and Copet, Jade and Synnaeve, Gabriel and Adi, Yossi},
  journal={arXiv preprint arXiv:2210.13438},
  year={2022}
}

@article{dac,
  title={High-fidelity audio compression with improved rvqgan},
  author={Kumar, Rithesh and Seetharaman, Prem and Luebs, Alejandro and Kumar, Ishaan and Kumar, Kundan},
  journal={Advances in Neural Information Processing Systems},
  volume={36},
  pages={27980--27993},
  year={2023}
}

@inproceedings{audiogen,
  title={Audiogen: Textually guided audio generation},
  author={Kreuk, Felix and Synnaeve, Gabriel and Polyak, Adam and Singer, Uriel and D{\'e}fossez, Alexandre and Copet, Jade and Parikh, Devi and Taigman, Yaniv and Adi, Yossi},
  booktitle={International Conference on Learning Representation},
  year={2023}
}

@article{audioX,
  title={Audiox: Diffusion transformer for anything-to-audio generation},
  author={Tian, Zeyue and Jin, Yizhu and Liu, Zhaoyang and Yuan, Ruibin and Tan, Xu and Chen, Qifeng and Xue, Wei and Guo, Yike},
  journal={arXiv preprint arXiv:2503.10522},
  year={2025}
}

@article{GenAU,
  title={Taming data and transformers for audio generation},
  author={Haji-Ali, Moayed and Menapace, Willi and Siarohin, Aliaksandr and Balakrishnan, Guha and Tulyakov, Sergey and Ordonez, Vicente},
  journal={arXiv preprint arXiv:2406.19388},
  year={2024}
}

@inproceedings{MMAudio,
  title={Taming multimodal joint training for high-quality video-to-audio synthesis},
  author={Cheng, Ho Kei and Ishii, Masato and Hayakawa, Akio and Shibuya, Takashi and Schwing, Alexander and Mitsufuji, Yuki},
  booktitle={Proceedings of the IEEE/CVF conference on computer vision and pattern recognition},
  year={2025}
}

@article{PPO,
  title={Proximal policy optimization algorithms},
  author={Schulman, John and Wolski, Filip and Dhariwal, Prafulla and Radford, Alec and Klimov, Oleg},
  journal={arXiv preprint arXiv:1707.06347},
  year={2017}
}

@inproceedings{CLAP,
  title={Large-scale contrastive language-audio pretraining with feature fusion and keyword-to-caption augmentation},
  author={Wu, Yusong and Chen, Ke and Zhang, Tianyu and Hui, Yuchen and Berg-Kirkpatrick, Taylor and Dubnov, Shlomo},
  booktitle={ICASSP 2023-2023 IEEE International Conference on Acoustics, Speech and Signal Processing (ICASSP)},
  pages={1--5},
  year={2023},
  organization={IEEE}
}

@inproceedings{FlowMatching,
  author       = {Yaron Lipman and
                  Ricky T. Q. Chen and
                  Heli Ben{-}Hamu and
                  Maximilian Nickel and
                  Matthew Le},
  title        = {Flow Matching for Generative Modeling},
  booktitle    = {International Conference on Learning Representations},
  year         = {2023},
}

@inproceedings{FMRecon,
  author       = {Peng Liu and
                  Dongyang Dai and
                  Zhiyong Wu},
  title        = {RFWave: Multi-band Rectified Flow for Audio Waveform Reconstruction},
  booktitle    = {International Conference on Learning Representations},
  year         = {2025},
 
}

@article{kong2020panns,
  title={Panns: Large-scale pretrained audio neural networks for audio pattern recognition},
  author={Kong, Qiuqiang and Cao, Yin and Iqbal, Turab and Wang, Yuxuan and Wang, Wenwu and Plumbley, Mark D},
  journal={IEEE/ACM Transactions on Audio, Speech, and Language Processing},
  volume={28},
  pages={2880--2894},
  year={2020},
  publisher={IEEE}
}

@article{kilgour2018fr,
  title={Fr$\backslash$'echet audio distance: A metric for evaluating music enhancement algorithms},
  author={Kilgour, Kevin and Zuluaga, Mauricio and Roblek, Dominik and Sharifi, Matthew},
  journal={arXiv preprint arXiv:1812.08466},
  year={2018}
}

@inproceedings{hershey2017cnn,
  title={CNN architectures for large-scale audio classification},
  author={Hershey, Shawn and Chaudhuri, Sourish and Ellis, Daniel PW and Gemmeke, Jort F and Jansen, Aren and Moore, R Channing and Plakal, Manoj and Platt, Devin and Saurous, Rif A and Seybold, Bryan and others},
  booktitle={2017 ieee international conference on acoustics, speech and signal processing (icassp)},
  pages={131--135},
  year={2017},
  organization={IEEE}
}

@article{li2022diffusion,
  title={Diffusion-lm improves controllable text generation},
  author={Li, Xiang and Thickstun, John and Gulrajani, Ishaan and Liang, Percy S and Hashimoto, Tatsunori B},
  journal={Advances in neural information processing systems},
  volume={35},
  pages={4328--4343},
  year={2022}
}

@inproceedings{hua-wang-2020-pair,
    title = "{PAIR}: Planning and Iterative Refinement in Pre-trained Transformers for Long Text Generation",
    author = "Hua, Xinyu  and
      Wang, Lu",
    editor = "Webber, Bonnie  and
      Cohn, Trevor  and
      He, Yulan  and
      Liu, Yang",
    booktitle = "Proceedings of the 2020 Conference on Empirical Methods in Natural Language Processing (EMNLP)",
    month = nov,
    year = "2020",
    address = "Online",
    publisher = "Association for Computational Linguistics",
    url = "https://aclanthology.org/2020.emnlp-main.57/",
    doi = "10.18653/v1/2020.emnlp-main.57",
    pages = "781--793"
}

@inproceedings{shen-etal-2019-towards,
    title = "Towards Generating Long and Coherent Text with Multi-Level Latent Variable Models",
    author = "Shen, Dinghan  and
      Celikyilmaz, Asli  and
      Zhang, Yizhe  and
      Chen, Liqun  and
      Wang, Xin  and
      Gao, Jianfeng  and
      Carin, Lawrence",
    editor = "Korhonen, Anna  and
      Traum, David  and
      M{\`a}rquez, Llu{\'i}s",
    booktitle = "Proceedings of the 57th Annual Meeting of the Association for Computational Linguistics",
    month = jul,
    year = "2019",
    address = "Florence, Italy",
    publisher = "Association for Computational Linguistics",
    url = "https://aclanthology.org/P19-1200/",
    doi = "10.18653/v1/P19-1200",
    pages = "2079--2089"
}

@inproceedings{hu-etal-2022-planet,
    title = "{PLANET}: Dynamic Content Planning in Autoregressive Transformers for Long-form Text Generation",
    author = "Hu, Zhe  and
      Chan, Hou Pong  and
      Liu, Jiachen  and
      Xiao, Xinyan  and
      Wu, Hua  and
      Huang, Lifu",
    editor = "Muresan, Smaranda  and
      Nakov, Preslav  and
      Villavicencio, Aline",
    booktitle = "Proceedings of the 60th Annual Meeting of the Association for Computational Linguistics (Volume 1: Long Papers)",
    month = may,
    year = "2022",
    address = "Dublin, Ireland",
    publisher = "Association for Computational Linguistics",
    url = "https://aclanthology.org/2022.acl-long.163/",
    doi = "10.18653/v1/2022.acl-long.163",
    pages = "2288--2305"
}

@article{hovy1990pragmatics,
  title={Pragmatics and natural language generation},
  author={Hovy, Eduard H},
  journal={Artificial Intelligence},
  volume={43},
  number={2},
  pages={153--197},
  year={1990},
  publisher={Elsevier}
}

@inproceedings{pan-mckeown-1998-learning-intonation,
    title = "Learning Intonation Rules for Concept to Speech Generation",
    author = "Pan, Shimei  and
      McKeown, Kathleen",
    booktitle = "36th Annual Meeting of the Association for Computational Linguistics and 17th International Conference on Computational Linguistics, Volume 2",
    month = aug,
    year = "1998",
    address = "Montreal, Quebec, Canada",
    publisher = "Association for Computational Linguistics",
    url = "https://aclanthology.org/P98-2165/",
    doi = "10.3115/980691.980734",
    pages = "1003--1009"
}

@article{mckeown1985discourse,
  title={Discourse strategies for generating natural-language text},
  author={McKeown, Kathleen R},
  journal={Artificial intelligence},
  volume={27},
  number={1},
  pages={1--41},
  year={1985},
  publisher={Elsevier}
}

@article{stiennon2020learning,
  title={Learning to summarize with human feedback},
  author={Stiennon, Nisan and Ouyang, Long and Wu, Jeffrey and Ziegler, Daniel and Lowe, Ryan and Voss, Chelsea and Radford, Alec and Amodei, Dario and Christiano, Paul F},
  journal={Advances in neural information processing systems},
  volume={33},
  pages={3008--3021},
  year={2020}
}

@article{snell2024scaling,
  title={Scaling llm test-time compute optimally can be more effective than scaling model parameters},
  author={Snell, Charlie and Lee, Jaehoon and Xu, Kelvin and Kumar, Aviral},
  journal={arXiv preprint arXiv:2408.03314},
  year={2024}
}

@article{ji2025survey,
  title={A Survey of Test-Time Compute: From Intuitive Inference to Deliberate Reasoning},
  author={Ji, Yixin and Li, Juntao and Xiang, Yang and Ye, Hai and Wu, Kaixin and Yao, Kai and Xu, Jia and Mo, Linjian and Zhang, Min},
  journal={arXiv preprint arXiv:2501.02497},
  year={2025}
}

@article{muennighoff2025s1,
  title={s1: Simple test-time scaling},
  author={Muennighoff, Niklas and Yang, Zitong and Shi, Weijia and Li, Xiang Lisa and Fei-Fei, Li and Hajishirzi, Hannaneh and Zettlemoyer, Luke and Liang, Percy and Cand{\`e}s, Emmanuel and Hashimoto, Tatsunori},
  journal={arXiv preprint arXiv:2501.19393},
  year={2025}
}

@article{wang2022self,
  title={Self-consistency improves chain of thought reasoning in language models},
  author={Wang, Xuezhi and Wei, Jason and Schuurmans, Dale and Le, Quoc and Chi, Ed and Narang, Sharan and Chowdhery, Aakanksha and Zhou, Denny},
  journal={arXiv preprint arXiv:2203.11171},
  year={2022}
}

@article{toh2024not,
  title={Not all votes count! programs as verifiers improve self-consistency of language models for math reasoning},
  author={Toh, Vernon YH and Ghosal, Deepanway and Poria, Soujanya},
  journal={arXiv preprint arXiv:2410.12608},
  year={2024}
}

@article{cobbe2021training,
  title={Training verifiers to solve math word problems},
  author={Cobbe, Karl and Kosaraju, Vineet and Bavarian, Mohammad and Chen, Mark and Jun, Heewoo and Kaiser, Lukasz and Plappert, Matthias and Tworek, Jerry and Hilton, Jacob and Nakano, Reiichiro and others},
  journal={arXiv preprint arXiv:2110.14168},
  year={2021}
}

@article{levi2024simple,
  title={A simple model of inference scaling laws},
  author={Levi, Noam},
  journal={arXiv preprint arXiv:2410.16377},
  year={2024}
}

@article{liu2024acemath,
  title={Acemath: Advancing frontier math reasoning with post-training and reward modeling},
  author={Liu, Zihan and Chen, Yang and Shoeybi, Mohammad and Catanzaro, Bryan and Ping, Wei},
  journal={arXiv preprint arXiv:2412.15084},
  year={2024}
}

@article{xin2024deepseek,
  title={Deepseek-prover: Advancing theorem proving in llms through large-scale synthetic data},
  author={Xin, Huajian and Guo, Daya and Shao, Zhihong and Ren, Zhizhou and Zhu, Qihao and Liu, Bo and Ruan, Chong and Li, Wenda and Liang, Xiaodan},
  journal={arXiv preprint arXiv:2405.14333},
  year={2024}
}

@article{ma2023let,
  title={Let's reward step by step: Step-Level reward model as the Navigators for Reasoning},
  author={Ma, Qianli and Zhou, Haotian and Liu, Tingkai and Yuan, Jianbo and Liu, Pengfei and You, Yang and Yang, Hongxia},
  journal={arXiv preprint arXiv:2310.10080},
  year={2023}
}

@article{khalifa2025process,
  title={Process reward models that think},
  author={Khalifa, Muhammad and Agarwal, Rishabh and Logeswaran, Lajanugen and Kim, Jaekyeom and Peng, Hao and Lee, Moontae and Lee, Honglak and Wang, Lu},
  journal={arXiv preprint arXiv:2504.16828},
  year={2025}
}

@inproceedings{
wang2025language,
title={Language Model Based Text-to-Audio Generation: Anti-Causally Aligned Collaborative Residual Transformers},
author={Juncheng Wang and Chao Xu and Cheng Yu and Zhe Hu and Haoyu Xie and Guoqi Yu and Lei Shang and Shujun Wang},
booktitle={The 2025 Conference on Empirical Methods in Natural Language Processing},
year={2025}
}

@article{comanici2025gemini,
  title={Gemini 2.5: Pushing the frontier with advanced reasoning, multimodality, long context, and next generation agentic capabilities},
  author={Comanici, Gheorghe and Bieber, Eric and Schaekermann, Mike and Pasupat, Ice and Sachdeva, Noveen and Dhillon, Inderjit and Blistein, Marcel and Ram, Ori and Zhang, Dan and Rosen, Evan and others},
  journal={arXiv preprint arXiv:2507.06261},
  year={2025}
}
\clearpage
\appendix
\section{Data Engine}
We detail our data pipeline for curating rollout data generated by a baseline audio synthesis model, using either synthesized or collected textual prompts.

\subsection{Synthesized Prompts}
To enrich the diversity and coverage of our training data, we employ a large language model (LLM)~\citep{comanici2025gemini} to automatically generate a wide range of textual prompts that describe various auditory scenes. 
These prompts are carefully designed to span diverse acoustic environments, sound sources, spatial configurations, and contextual scenarios. The LLM is guided by a set of curated seed phrases and constraints to ensure semantic coherence, acoustic plausibility, and alignment with the capabilities of the audio generation model. 
This synthetic prompt generation process significantly expands the breadth of our dataset while maintaining control over the thematic and structural properties of the resulting audio scenes.

\subsection{Realistic Prompts Collection}
To mitigate the domain gap between synthetically generated prompts and naturally occurring audio descriptions, we curate a set of realistic prompts from the official test sets of two established audio-captioning benchmarks: AudioCaps~\citep{kim2019audiocaps} and VGGSound~\citep{chen2020vggsound}. These datasets provide realistic audio collected from real world, that reflect authentic listening experiences and real-world acoustic contexts.

To ensure a fair and standardized evaluation, we reserve 1,000 prompts from each dataset—drawn exclusively from their official test splits—as our held-out evaluation set. The remaining prompts from the test sets (where permitted by the dataset licenses and evaluation protocols) are repurposed to augment our training data, thereby enriching the realism and linguistic diversity of our prompt distribution without compromising the integrity of the final benchmark results. This strategy allows us to bridge the gap between synthetic and real-world prompts while maintaining compatibility with established evaluation frameworks.

\subsection{Rollout Data Collection}
Using the combined set of synthesized and realistic prompts described above, we generate rollout data with the pre-trained autoregressive audio generator \textit{Siren}~\citep{wang2025language}. For each prompt, we perform random sampling to produce 32 distinct audio rollouts, resulting in a diverse set of acoustic realizations conditioned on the same textual instruction.

For every generated audio–prompt pair, we compute an instruction-following score using the CLAP model~\citep{CLAP}. Specifically, CLAP provides a similarity-based alignment score between the generated audio and its corresponding prompt, which we interpret as a proxy for how well the audio adheres to the semantic and contextual intent of the instruction. These scores are assigned to each rollout trajectory, enabling us to assess and later rank or filter generations based on their fidelity to the given prompt. This annotated dataset of (prompt, audio, CLAP score) triples forms the foundation for subsequent critic optimization stages in our pipeline.

\section{Metrics Computation}
The primary objective evaluation metrics we employ are Fréchet Distance (FD), Inception Score (IS), and Kullback–Leibler (KL) divergence, all computed using features extracted from PANNs (Pretrained Audio Neural Networks)~\citep{kong2020panns}, a state-of-the-art audio classification model.
\begin{itemize}
    \item \textbf{Fréchet Distance} (FD), analogous to the Fréchet Inception Distance (FID) in image generation, quantifies the similarity between the distributions of generated and reference (real) audio samples in the PANNs embedding space. A lower FD indicates better alignment between the generated and target audio distributions.
    \item  \textbf{Inception Score} (IS) assesses both the quality and diversity of generated samples by measuring the entropy of conditional label distributions predicted by PANNs: high-quality samples yield confident predictions (low conditional entropy), while diverse samples produce a broad range of labels (high marginal entropy).
    \item  \textbf{KL divergence} is computed at the individual (paired) sample level—comparing the PANNs-predicted class distribution of a generated audio clip against that of its corresponding ground-truth or reference audio—and then averaged across the dataset to yield a final score. Lower KL divergence reflects closer semantic and acoustic alignment.
\end{itemize}

In addition to PANNs-based metrics, we also report the **Fréchet Audio Distance (FAD)**~\citep{kilgour2018fr}, which follows a similar distributional comparison principle but uses embeddings from VGGish~\citep{hershey2017cnn}. While VGGish is a widely used audio representation model, it is generally considered less powerful than PANNs for modern audio understanding tasks; thus, FAD serves as a complementary, albeit potentially less discriminative, benchmark.

Finally, to evaluate semantic alignment between audio and text, we adopt the CLAP score~\citep{CLAP}, which leverages the contrastively trained CLAP model to compute the cosine similarity between the embedded representations of generated audio and their associated prompts. Higher CLAP scores indicate stronger cross-modal coherence and better instruction following.

\section{Additional Ablate Study on Difference Sampling Strategies}
\begin{figure}
    \centering
    \includegraphics[width=.6\linewidth]{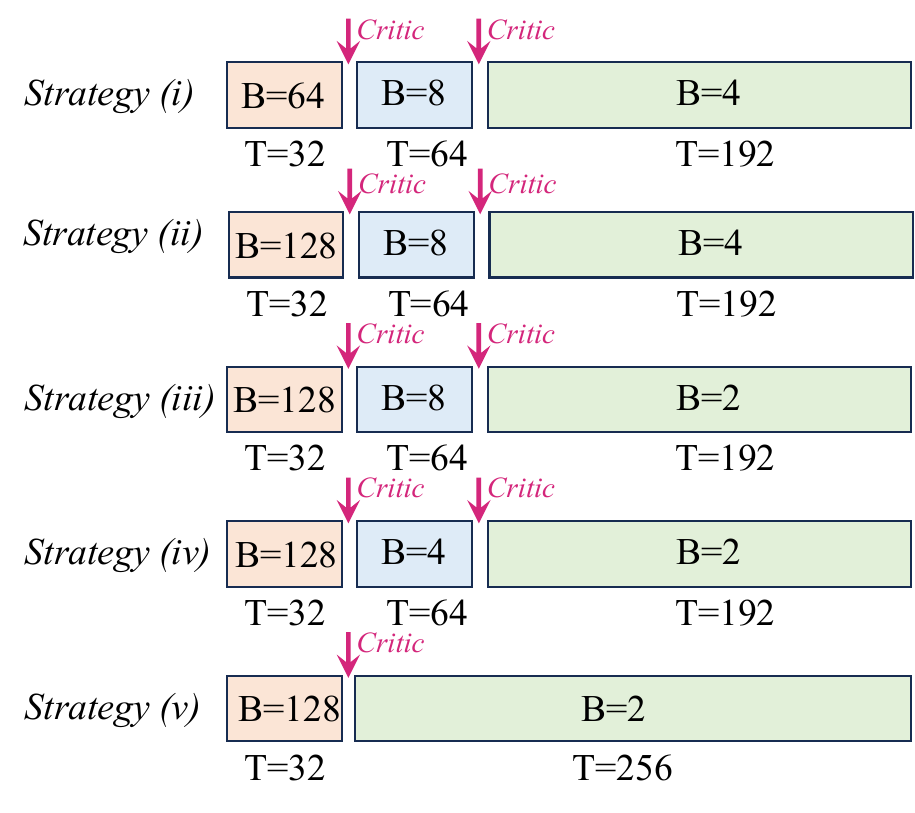}
    \caption{Schematic illustration of different sampling strategies, where $B$ is batch size, $T$ the temporal length. }
    \label{fig-strategy}
\end{figure}

In Table~\ref{tab-ablate_strategy}, we compare different kinds of sampling strategies with a fixed critic step of $T_{\text{prefix}} = 32$, where the schematic illustration of each strategy can be found in Figure~\ref{fig-strategy}.
% \usepackage[normalem]{ulem}
% \useunder\underline{ine}\underline{}{}
\begin{table}[]
\centering

\renewcommand{\arraystretch}{1.15}{
\resizebox{.5\textwidth}{!}{
\begin{tabular}{c|c|c|c|c|c|c}
\whline
\textbf{Sampling Strategy} & \textbf{Cost Tokens$\downarrow$} & \textbf{CLAP$\uparrow$} & FAD$\downarrow$ & FD$\downarrow$ & IS$\uparrow$ & KL$\downarrow$ \\ \whline
 
Strategy (i)               & 3456                             & 32.46                   & 1.87            & 16.30          & 12.00        & 1.50           \\ \hline
Strategy (ii)              & 5504                             & 36.98                   & 1.96            & 15.79          & 13.51        & 1.30           \\ \hline
Strategy (iii)             & 5056                             & 36.75                   & 1.94            & 15.75          & 13.56        & 1.29           \\ \hline
Strategy (iv)              & 4800                             & 36.74                   & 1.93            & 15.74          & 13.57        & 1.30           \\ \hline
\rowcolor[HTML]{EFEFEF}Strategy (v)               & 4608                             & 36.47                   & 1.96            & 15.70          & 13.82        & 1.27           \\ \whline
\end{tabular}
}}
\caption{Ablating different sampling strategies.}
\label{tab-ablate_strategy}
\end{table}

Strategy (i) generates 64 prefixes and completes all of them, resulting in modest instruction-following performance (CLAP: 32.46). Strategy (ii) explores a larger pool of 128 prefixes and prunes to the top 8 for completion, yielding a significant CLAP improvement (36.98) at a higher token cost (5,504), which exceeds our budget constraint. 

Strategies (iii)–(v) operate within the 4,608-token budget. Strategy (iii) mirrors our main method: 128 prefixes are generated, scored by Plan-Critic at step 32, and only the top 2 are completed, achieving a CLAP score of 36.75. Strategy (iv) retains 4 completions instead of 2, achieving comparable performance (CLAP: 36.74) with slightly higher cost (4,800 tokens). Strategy (v) serves as a strong baseline: it generates 128 prefixes and directly completes the top 2 \emph{without} an intermediate critique step, attaining CLAP 36.47—matching the result reported in Table~\ref{tab-main_result_AC} for our full method.

These results highlight two key insights: (1) early evaluation via Plan-Critic enables more effective pruning than selecting prefixes based on generation likelihood alone, and (2) allocating more computation to diverse prefix exploration—followed by aggressive pruning—yields substantial gains in semantic fidelity without increasing total inference cost. Our chosen configuration (Strategy~iii) thus offers the best trade-off between performance and efficiency.

\end{document}